\documentclass[lettersize,journal]{IEEEtran}
\usepackage{amsmath,amsfonts}
\usepackage{algorithmic}
\usepackage{array}
\usepackage[caption=false,font=normalsize,labelfont=sf,textfont=sf]{subfig}
\usepackage{textcomp}
\usepackage{stfloats}
\usepackage{url}
\usepackage{verbatim}
\usepackage{graphicx}
\usepackage{orcidlink}
\usepackage{mathtools}          
\usepackage[algo2e,ruled,vlined,linesnumbered]{algorithm2e}
\usepackage{xcolor}
\usepackage{booktabs}
\usepackage[noabbrev,nameinlink]{cleveref} 
\usepackage{adjustbox}
\usepackage{colortbl}           
\usepackage{multirow}

\def\BibTeX{{\rm B\kern-.05em{\sc i\kern-.025em b}\kern-.08em
    T\kern-.1667em\lower.7ex\hbox{E}\kern-.125emX}}
\usepackage{balance}

\DeclareMathOperator*{\argmax}{arg\,max}

\definecolor{pastelred}{rgb}{1.0, 0.6, 0.6}
\definecolor{pastelgreen}{rgb}{0.6, 1.0, 0.6}
\definecolor{pastelblue}{rgb}{0.6, 0.6, 1.0}
\definecolor{pastelcyan}{rgb}{0.6, 1.0, 1.0}
\definecolor{pastelmagenta}{rgb}{1.0, 0.6, 1.0}
\definecolor{pastelyellow}{rgb}{1.0, 1.0, 0.6}
\definecolor{pastelorange}{rgb}{1.0, 0.8, 0.6}
\definecolor{combinatorialcolor}{HTML}{708ec0}
\definecolor{factoredcolor}{HTML}{e49e75}

\definecolor{ppo_sp}{HTML}{3373a1}   
\definecolor{ppo_mp}{HTML}{e1812b}   
\definecolor{ddqn_mp}{HTML}{3b923a}  
\definecolor{theory}{HTML}{c03d3e}   

\newcommand{\irace}{\textsc{irace}\xspace}
\newcommand{\onell}{(1+($\lambda$,$\lambda$))-GA\xspace}
\newcommand{\onemax}{\textsc{OneMax}\xspace}

\newcommand{\algocmt}[1]{\hfill\textcolor{lightgray}{$\triangleright$ #1}} 

\DeclareMathOperator{\flip}{flip}
\DeclareMathOperator{\cross}{cross}

\providecommand{\orcidID}[1]{}

\begin{document}

\title{Discovering Interpretable Multi-Parameter Control Policies for Evolutionary Algorithms Using Deep Reinforcement Learning}   





\author{%
     Tai Nguyen\orcidlink{0009-0004-7707-2069},%
    \thanks{Tai Nguyen is with University of St Andrews, St Andrews, United Kingdom and Sorbonne Université, CNRS, LIP6, Paris, France (Email: {\tt dptn1@st-andrews.ac.uk}).} %
    Phong Le\orcidlink{0009-0000-0749-9519}, %
     \thanks{Phong Le is with University of St Andrews, St Andrews, United Kingdom (Email: {\tt pl200@st-andrews.ac.uk}).}%
    Carola Doerr\orcidlink{0000-0002-4981-3227}, %
    \thanks{Carola Doerr is with Sorbonne Université, CNRS, LIP6, Paris, France (Email: {\tt carola.doerr@lip6.fr}).}%
    Nguyen Dang\orcidlink{0000-0002-2693-6953} %
   \thanks{Nguyen Dang is with University of St Andrews, St Andrews, United Kingdom (Email: {\tt nttd@st-andrews.ac.uk}).}%
}

\markboth{~}%
{Discovering the Multi-parameter Control Policies of the (1+($\lambda$,$\lambda$))-GA via Deep Reinforcement Learning}

\maketitle

\begin{abstract}

While deep Reinforcement Learning (deep-RL) has been increasingly applied to parameter control in evolutionary algorithms, rigorous theoretical analysis of parameter control remains largely restricted to single-parameter settings, owing to the difficulty of deriving effective, interpretable multi-parameter policies amenable to formal study. We demonstrate how deep-RL can be leveraged to overcome this barrier, using the (1+($\lambda$,$\lambda$))-genetic algorithm optimizing OneMax, one of the few problems where a super-constant speedup of dynamic control has been formally proven, as a representative case study. We first show that standard approaches struggle to converge in this multi-parameter setting, and introduce algorithm-agnostic enhancements targeting action-space decomposition, reward shifting, and long-horizon discounting. With these in place, we compare common deep-RL methods and find that Double Deep Q-Networks uniquely avoid the policy collapse observed in Proximal Policy Optimization, yielding trajectories suitable for downstream analysis. Crucially, we move beyond the ``black-box'' nature of neural networks by distilling the learned behaviors into a transparent, symbolic control policy. This resulting policy does not only offer interpretability for future theoretical analysis but also yields exceptional performance, consistently outperforming existing baselines across a wide range of problem sizes.

\end{abstract}


\section{Introduction}

\IEEEPARstart{P}{arameter control} studies how to best adjust the parameters of an evolutionary algorithm (EA) during its execution~\cite{eiben2007parameter}. State-of-the-art EAs often strongly benefit from parameter control methods~\cite{aleti2016systematic}, as these methods enable the algorithms to adapt their search behavior to meet the varying requirements of different stages in the optimization process. For example, it is often desirable to converge from a global search exploring the most promising regions of the search space to a local search, which exploits the latter to identify local optima. Parameter control is an intensively studied topic in evolutionary computation (EC), both from an empirical angle~\cite{KarafotiasSE12,EibenHM99,AletiM16} and from a running time analysis perspective~\cite{doerr2020theory}. However, identifying optimal or reasonably good control policies remains a challenging endeavor. 

Traditionally, identifying these parameter control policies has relied on paradigms with inherent limitations. Deterministic schedules are typically inflexible to handle complex search landscapes, while hand-crafted adaptive policies~\cite{hansen1996adapting,auger2005restart} demand deep, domain-specific expert intuition. Conversely, self-adaptive mechanisms attempt to evolve the parameters alongside the candidate solutions~\cite{karafotias2014parameter}, but they often incur substantial computational overhead~\cite{bi2011classification}, delayed feedback, and potential misalignment between parameter values and fitness gains~\cite{qin2005self,mallipeddi2008empirical}. Another popular direction involves ``on-the-fly'' online learning via bandit-based approaches~\cite{dacosta2008adaptive,fialho2010analyzing,DoerrDY16ppsn}; however, because these methods learn from scratch during each run, they fundamentally lack the capacity to transfer learned algorithmic behaviors across different problem instances.

\begin{figure}[t]
    \centering
    \includegraphics[width=\linewidth, clip]{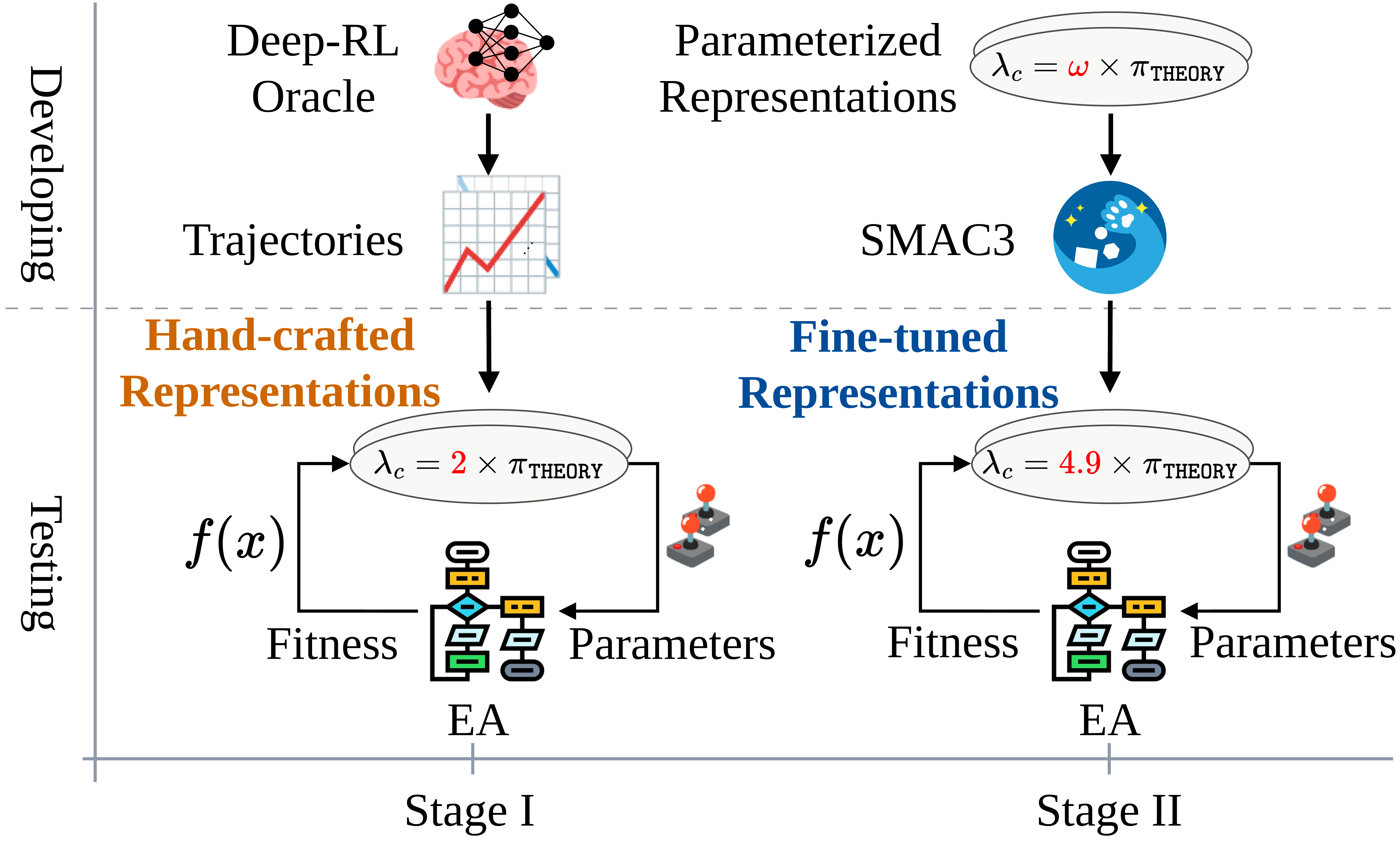}
    \caption{\textbf{The proposed two-stage distillation framework for discovering symbolic multi-parameter control policies.} The DAC setting of \onell solving the \onemax problem is represented by the loop (bottom). To bridge the gap between empirical deep-RL performance and theoretical interpretability, our methodology operates in two stages. \textbf{Stage I:} A deep-RL oracle generates optimal parameter trajectories, which are distilled into explicit, \emph{Hand-crafted Representations} (e.g., $\lambda_c = 2 \times \pi_{\text{THEORY}}$). \textbf{Stage II:} These equations are generalized into \emph{Parameterized Representations} (e.g., $\lambda_c = \omega \times \pi_{\text{THEORY}}$) and optimized via SMAC3~\cite{lindauer2022smac3}, resulting in \emph{Fine-tuned Representations} (e.g., $\lambda_c = 4.9 \times \pi_{\text{THEORY}}$).}
    \label{fig:tevc_framework}
\end{figure}

To overcome barriers of conventional parameter control methodologies, the EC community has increasingly turned to the \emph{Dynamic Algorithm Configuration} (DAC) paradigm, where deep reinforcement learning (deep-RL) has emerged as the state-of-the-art methodology for parameter control~\cite{sharma2019deep,shala2020learning,sun2021learning}. In this framework, the algorithm-problem interaction is modeled sequentially: the optimization progress is characterized by observable \emph{states}, and the EA parameters serve as the \emph{actions}. By interacting with the environment through a dedicated ``offline'' training phase, an RL agent learns a robust policy that maps states to hopefully optimal parameter adjustments or actions. Crucially, these learned policies can then be deployed online on unseen problem instances, successfully enabling the transfer of effective parameter trajectories, an advantage never achieved in bandit-strategy approaches.

Although recent studies demonstrate that deep-RL agents can discover policies that drastically outperform traditional baselines, relying entirely on these neural networks creates a critical ``black-box'' bottleneck for both theoreticians and practitioners. To truly advance algorithmic configuration, they do not just need a black-box oracle that outputs good parameter values; they need a \emph{symbolic blueprint}, explicit mathematical rules that can be analyzed, verified, and generalized. Specifically, transitioning from neural to symbolic representation addresses several core limitations in modern DAC. First, unlike neural networks that often overfit to distributions of training instances~\cite{sharma2019deep,ma2024auto}, mathematical operators (i.e., building blocks of symbolic representations) are inherently agnostic to problem-specific details. A symbolic update rule operates identically regardless of the problem dimension, enabling zero-shot generalization across entirely unseen problem instances and extended optimization horizons. From a computational and theoretical perspective, symbolic policies also offer stark advantages in efficiency and transparency. Evaluating a simple algebraic equation incurs merely an $\mathcal{O}(1)$ overhead, bypassing the heavy matrix multiplications of deep network forward passes. Most crucially, symbolic representations provide the actionable interpretability required for formal running time analysis. While the distributed weights of a neural network fundamentally obscure the precise rules driving algorithmic phase transitions, symbolic approaches distill these strategies into explicit, human-readable equations. This allows researchers to transparently observe algorithmic phase transitions, directly bridging the gap between empirical policy discovery and theoretical verification.

This work is an extension of~\cite{nguyen2025multi} that employs Double Deep Q-Networks (DDQN), a widely used deep-RL method, to uncover effective hand-crafted symbolic representations for multi-parameter control policies in the $(1+(\lambda,\lambda))$-GA on \onemax. Building on the initial successes of that earlier work, we continue to bridge the gap between empirical deep-RL and theoretical EC by offering diverse analyses of various deep-RL methods and then distilling the symbolic policies in a two-stage process. Our contributions are: 
    
\begin{itemize}
\item \textbf{Comparative efficacy of deep-RL methods:} To identify the most robust deep-RL oracle for distilling the symbolic representations, we conduct a comprehensive empirical comparison between two commonly used algorithms: DDQN (expanding the results from~\cite{nguyen2025multi}) and Proximal Policy Optimization (PPO). More importantly, we conduct an extensive experiment on reward shifting in PPO, which is crucial for the effectiveness of DDQN in this context~\cite{nguyen2025importance}; however, its applicability to PPO remains an open question. We observe that, while PPO is widely praised for its stability in general RL environments, such as games and robotics, it frequently experiences policy collapse due to the challenging exploration required in this specific DAC environment. Conversely, the DDQN agent from~\cite{nguyen2025multi} successfully navigates this challenge, cementing its role as the effective deep-RL algorithm to generate the high-quality trajectories required for our symbolic extraction.
\item \textbf{Discovery of symbolic policies:} We elevate the symbolic distillation proposed in~\cite{nguyen2025multi} to a new level by proposing a two-stage process to derive simple, interpretable, yet highly effective multi-parameter control policies, as shown in~\Cref{fig:tevc_framework}. We consider the hand-crafted symbolic equations, informed by the behavior of high-performing deep-RL agents in~\cite{nguyen2025multi}, as the initial stage. Subsequently, we refine these mathematical structures through automated algorithm configuration, yielding a more powerful symbolic policy that surpasses the original formulation in~\cite{nguyen2025multi} by $6.8\%$.

\item \textbf{Bridging theory and practice:} Our derived symbolic policies significantly outperform previously known parameter control rules. By providing explicit mathematical control equations, we supply theory-focused researchers with formulas that can stimulate the extension and refinement of existing runtime proofs. Simultaneously, we highlight critical design considerations for empirical researchers employing RL in algorithmic contexts.
\end{itemize}

\textbf{Reproducibility:} Our code and data are available at~\cite{githubsource}. 

\section{Background}
\label{sec:background}
This section offers the necessary background for this work. We first review the Dynamic Algorithm Configuration problem, the natural alignment between the problem and the sequential decision-making problem targeted by RL. Finally, we briefly describe the two commonly used deep-RL algorithms, namely Deep Q-networks and Proximal Policy Optimization, which are used in this study.

\subsection{Dynamic Algorithm Configuration (DAC)}
\label{sec:backgroundDAC}


The task of automated algorithm configuration aims at finding the best parameter setting(s) for a given parameterized algorithm. This task can be categorized into two main paradigms: \emph{static} and \emph{dynamic}~\cite{eiben1999parameter,birattari2009tuning,biedenkapp2020dynamic}. The static algorithm configuration task, which is often referred to as \emph{algorithm configuration (AC)}, aims at finding a parameter setting that remains constant throughout the algorithm's execution. More formally, it involves determining a parameter setting (i.e., a configuration) $\theta^* \in \Theta$ for a specific algorithm $A$ that achieves the best possible expected performance with respect to an instance distribution $\mathcal{D}_\mathcal{I}$  over a set $\mathcal{I}$ of problem instances~\cite{hutter2009paramils}.
Here, the performance of an algorithm configuration $\theta$ on a given problem instance $i$ is defined by the cost function $c(\theta,i)$ (e.g., the average runtime to solve the instance). Formally, the AC problem aims at finding the optimal configuration 
$\theta^* \in \arg\min_{\theta \in \Theta} \mathbb{E}_{i \sim \mathcal{D}_\mathcal{I}} c(\theta,i)$.


On the other hand, \emph{dynamic algorithm configuration (DAC)} aims at dynamically adjusting algorithm parameters \emph{during} the solving process~\cite{biedenkapp2020dynamic}. More formally, DAC involves learning a parameter control policy $\pi^{*}: \mathcal{S} \times \mathcal{I} \rightarrow \Theta$, where $\pi^{*}(s,i)$ is a mapping function from the current state $s$ of the solving process on a problem instance $i$ to the best parameter setting $\theta^* \in \Theta$ (i.e., configuration) for that pair of state and instance. In other words, 
$\pi^{*} \in \arg\min_{\pi} \mathbb{E}_{i \sim \mathcal{D}_\mathcal{I}}c(\pi,i)$, 
where $c(\pi, i)$ is the cost of solving instance $i$ using the given algorithm $A$ under the parameter control policy $\pi$.


There is a large body of existing works in the literature that address similar dynamic algorithm configuration settings. They include \emph{parameter control}~\cite{eiben1999parameter}, \emph{operator selection}~\cite{pettinger2002controlling}, \emph{automated algorithm selection}~\cite{lagoudakis2000algorithm}, and  \emph{hyper-heuristics}~\cite{burke2013hyper,qu2020general}. In this study, we adopt the formal definition proposed in Biedenkapp et al.~\cite{biedenkapp2020dynamic} and use the term DAC to refer to the considered learning setting, where the parameter control policy is represented as a machine learning model (in our case, a neural network) and is learned via a dedicated \emph{offline} training phase. A natural learning framework for this training phase, as we describe in the next section, is deep-RL.

\subsection{Deep Reinforcement Learning}
\label{sec:backgroundRL}

As proposed in~\cite{biedenkapp2020dynamic}, DAC can be formulated as a contextual Markov Decision Process (MDP) $\mathcal{M}_\mathcal{I}:= \{ \mathcal{M}_i\}_{i\in\mathcal{I}}$. For a given problem instance $i \in \mathcal{I}$, the contextual MDP $\mathcal{M}_{i}$ is given by $\left< \mathcal{S},\mathcal{A},\mathcal{T}_i,\mathcal{R}_i \right>$, where $\mathcal{S}$ represents the state space (e.g., a cost metric of the configured target algorithm $A$), $\mathcal{A}$ is the action space (i.e,. used to adjust configuration $\theta$), $\mathcal{T}_i$ is the transition function $\mathcal{S}\times\mathcal{A}\times\mathcal{S} \rightarrow [ 0,1 ]$ that maps a triple $(s, a, s')$ to the probability $\mathcal{T}_i(s, a, s') = P(s_{t+1} = s' \mid s_t = s, a_t = a)$ that choosing action $a$ in state $s$ leads to state $s'$, and the reward $\mathcal{R}_i: \mathcal{S}\times\mathcal{A} \rightarrow \mathbb{R}$ is the reward after the transition that is intended to be designed to reflect the quality of applying action $a$ when observing state $s_t$.

By formulating DAC within the contextual MDP framework, reinforcement learning, a class of solutions for MDP, is clearly well-suited for data-driven DAC~\cite{sharma2019deep,biedenkapp2020dynamic}. The application of RL for DAC can be divided into \emph{offline} and \emph{online} DAC policies~\cite{adriaensen2022automated}, distinguished by the timing of policy execution for controlling the configuration (i.e., use time). In an offline setting, a policy is trained using a training set and then applies the acquired knowledge to solve a new or unseen testing set. Conversely, in the online DAC setting, there is no concept of ``use time,'' as the policy continuously learns and solves the problem simultaneously. The diverse successes of RL for DAC have been prominently observed in offline setting due to its generalization~\cite{tessari2022reinforcement,speck2021learning,nguyen2025deep,xu2024accelerate,guo2025reinforcement}. During the offline configuration phase, the RL policy engages in \emph{trial-and-error} interactions with the environment (i.e., the target algorithm $A$ solving a set of problem instances). At each timestep $t$, the RL policy selects an action $a_t$ based on its estimation of the action's quality. It begins by observing the state information $s_t$, recalling its learned knowledge, and then making a decision. Immediately, the RL agent receives a reward signal $r_t$ measuring the true quality of the selected action, uses that reward to update its policy, and then the environment transitions to a new state $s_{t+1}$. Typically, the RL policy is stored in a matrix with dimensions $|\mathcal{S}| \times |\mathcal{A}|$, where each row corresponds to a unique state $s$ and each column corresponds to a distinct action $a$. Each cell $(s, a)$ holds the current estimate of the expected return (i.e., cumulative reward) for that particular state-action pair. The training objective is to maximize the return until the end of a running episode. For problems with small-moderate state and action spaces, a look-up table (i.e., matrix) is often utilized to keep track the policy, commonly known as a $\mathcal{Q}$-table. However, as the dimensionality of these spaces increases, the look-up table approach becomes obsolete and is surpassed by robust function approximation techniques employing neural networks, commonly referred to as \emph{deep}-RL.

\textbf{Double Deep Q-Network (DDQN).} One of the most widely used deep-RL approaches is DDQN~\cite{van2016deep}, which employs a neural network to approximate the action-value function with parameters $\phi$, denoted as $\mathcal{Q}(s, a; \phi)$. The DDQN represents an enhancement of the original Deep Q-Network (DQN)~\cite{mnih2015human}, wherein the policy training process is designed to learn a $\mathcal{Q}$-function (i.e., value function) that quantifies the expected quality of a particular action given a specific state. This objective introduced by Watkins et al.~\cite{watkins1992q}, employs a temporal-difference (TD) method to update the $\mathcal{Q}$-function using the immediate reward $r_t$ observed following the agent’s action. The update is defined as:
\begin{equation*}
\text{TD}(s_t, a_t;\phi) = r_t + \gamma \max_a \mathcal{Q}(s_{t+1}, a;\phi) - \mathcal{Q}(s_t, a_t;\phi),
\end{equation*}
where $\gamma$ denotes the discount factor, which regulates the significance assigned to future rewards in the estimation process. Unlike the traditional DQN, DDQN employs two distinct $\mathcal{Q}$-functions to separate the steps of selecting and evaluating the optimal next action, thereby reducing overestimation bias. The policy learned by DDQN can be expressed as: $\pi(s) = \argmax_a  \mathcal{Q}(s, a;\phi)$. DDQN is widely recognized as a prominent representative of value-based RL methods, noted for their improved sample efficiency. In value-based RL, a replay buffer is employed to store transitions generated by the policy over time. During the training phase, the agent samples batches of transitions from this buffer, allowing it to learn effectively from successful experiences generated by previous policies while mitigating correlations between consecutive samples. This mechanism enhances learning stability and overall performance.

\textbf{Proximal Policy Optimization (PPO).} Policy gradient methods directly parameterize the control policy $\pi(a|s;\phi)$ unlike learning via a value function such as DQN. While early approaches like REINFORCE~\cite{williams1992simple} utilize the policy gradient theorem to optimize parameters, they suffered from high variance and sample inefficiency due to their reliance on full trajectory returns. To mitigate this, modern architectures adopt an actor-critic framework~\cite{712192}, introducing a value function baseline to compute \emph{advantage function} $A(s, a) = \mathcal{Q}(s, a) - V(s)$. This metric isolates the specific contribution of an action relative to the state's average value, significantly stabilizing the learning signal. PPO~\cite{schulman2017proximal} refines the actor-critic approach by addressing a critical instability in standard policy gradients: the risk of destructively large policy updates. PPO enforces a trust region constraint not via complex second-order optimization, but through a simpler capped surrogate objective:
\begin{equation*}
\mathcal{L}(\phi) = \mathbb{E}_t \bigg [ \min\left( \rho_t(\phi) \hat{A}_t, \text{clip}(\rho_t(\phi), 1-\varepsilon, 1+\varepsilon) \hat{A}_t \right) \bigg ],
\end{equation*}
where the probability ratio $\rho_t(\phi) = \frac{\pi(a_t|s_t; \phi)}{\pi(a_t|s_t; \phi_{\text{old}})}$ quantifies the divergence between the current and the previous policies. The hyperparameter $\varepsilon$ specifies an acceptable range for policy updates. Concretely, it prevents large policy updates by clipping the objective when the new policy deviates too far from the old one. This limits the benefit of extreme changes, encouraging stable, incremental improvements rather than drastic updates.

\subsection{Our Use-Case: Controlling Four Parameters of the \texorpdfstring{\onell}{(1+(L,L)) GA} Optimizing \texorpdfstring{\onemax}{OneMax}}
\label{sec:backgroundga}

We choose a use case for which a super-constant advantage of dynamic parameter choices over any possible static ones has been formally proven, the optimization of the \onemax function using the $(1+(\lambda,\lambda))$~GA.

\paragraph{The $(1+(\lambda,\lambda))$~GA} 
We use a variant of the $(1+(\lambda,\lambda))$~GA with four controllable parameters: the \emph{population sizes} $\lambda_m ,\lambda_c \in \mathbb{N}$ for the mutation and crossover phases, respectively, the \emph{mutation rate coefficient} $\alpha\in \mathbb{R}_{>0}$, and the \emph{crossover bias coefficient} $\beta \in \mathbb{R}_{>0}$. In line with previous works~\cite{nguyen2025importance, nguyen2025deep, lissovoi2020simple, biedenkapp2022theory}, we seek control policies that use fitness-dependent parameter choices, i.e., we search policies that choose $\lambda_m$, $\lambda_c$, $\alpha$, and $\beta$ in dependence of the quality $f(x)$ of the current-best solution $x$. 

\begin{algorithm2e}[t]%
\caption{The $(1+(\lambda,\lambda))$-GA with four fitness-dependent parameters: the population sizes $\lambda_m$ (for the mutation phase) and $\lambda_c$ (for the crossover phase), the mutation rate coefficient $\alpha$, and the crossover bias coefficient $\beta$. \\\textbf{Input:} Problem size $n$, function $f: \{0,1\}^n \rightarrow \mathbb{R}$ }
\label{alg:onell}
$x \gets$ a sample from $\{0,1\}^{n}$ chosen u.a.r.\;
\While{$x$ is not the optimal solution}{

Choose $\lambda_m$, $\lambda_c$, $\alpha$ and $\beta$ based on $f(x)$ \\

\underline{\textbf{Mutation phase:}}\\
        $p = \alpha \lambda_m/n$; \hfill \algocmt{mutation rate} \\
	Sample $\ell$ from $\text{Bin}_{>0}(n,p)$\;
	\lFor{$i=1, \ldots, \lambda_m$}
         {$x^{(i)} \leftarrow \text{flip}_{\ell}(x)$; Evaluate $f(x^{(i)})$}
	Choose $x' \in \{x^{(1)}, \ldots, x^{(\lambda_m)}\}$ with $f(x')=\max\{f(x^{(1)}), \ldots, f(x^{(\lambda_m)})\}$ u.a.r.\;
\underline{\textbf{Crossover phase:}}\\
 $c = \beta/\lambda_c$; \hfill \algocmt{crossover bias} \\
 $\mathcal{Y} \gets \emptyset$\;
\For{$i=1, \ldots, \lambda_c$}
{$y^{(i)} \leftarrow \text{cross}_{c}(x,x')$\; 
\lIf{$y^{(i)} \notin \{x,x'\}$}{Evaluate $f(y^{(i)})$; Add $y^{(i)}$ to $\mathcal{Y}$}}
\underline{\textbf{Selection and update step:}}\\
Choose $y \in \{x'\}\cup\mathcal{Y}$ with 
    $f(y) = \max\{f(x'), \max_{y' \in \mathcal{Y}}f(y')\}$ u.a.r.\;
\lIf{$f(y)\ge f(x)$}{$x \leftarrow y$} 
}
\end{algorithm2e}

Algorithm~\ref{alg:onell} summarizes this algorithm, which works as follows. Starting from an initial solution, the \onell iteratively runs until an optimal solution is found.\footnote{In practice one would cap the maximally allowed number of evaluations. However, here we focus on investigating (and eventually minimizing) the expected optimization time and therefore ignore this implementation detail by choosing a large enough evaluation budget that allows the \onell to find the optimum with a very high probability.} 
In each iteration, the algorithm consists of three phases: the mutation phase, the crossover phase, and the selection step. 

\textbf{Mutation phase.} 
During the mutation phase, $\lambda_m$ offspring solutions are created by applying standard bit mutation to the current-best solution $x$. More precisely, a mutation strength $\ell$ is sampled from the binomial distribution $\text{Bin}_{>0}(n,p)$ that is conditioned on returning a value $\ell>0$ through rejection sampling; i.e., $\ell$ is sampled from $\text{Bin}(n,p)$ until a positive value is returned. Following the suggestions made in~\cite{doerr2015black,doerr2018optimal}, the \emph{mutation rate} $p\in (0, 1]$ is parametrized as $p=\alpha \lambda_m / n$. Each of the $\lambda_m$ offspring $x^{(i)}$ is then obtained by applying the $\flip_{\ell}$ operator to $x$, which flips exactly $\ell$ pairwise different bits in $x$, chosen uniformly at random (u.a.r.).  

\textbf{Crossover phase.} 
The top solution among the $\lambda_m$ offspring obtained during the mutation phase, labeled $x'$, is chosen for the crossover phase; in case of ties a random one among the best offspring is chosen. It is worth noting that $x'$ may (and frequently will) be worse than the current incumbent solution $x$. The goal of the crossover phase is to \emph{repair} it, see~\cite{doerr2015black} for a detailed intuition behind the \onell. In a nutshell, the crossover acts as a \emph{genetic repair mechanism} that mitigates the potential loss of beneficial genes from the parent due to the \emph{aggressive} mutation rate while preserving them in the offspring. To this end, $\lambda_c$ offspring are generated by combining $x'$ with its parent $x$ through the biased crossover operator $\cross_c$, which creates a new search point by choosing, independently for each position $1 \le i \le n$, the entry of the second input with probability $c$ and choosing the entry from the first argument otherwise. We refer to $c\in (0,1]$ as the \emph{crossover bias,} and we parametrize $c=\beta/\lambda_c$, again following the suggestions made in~\cite{doerr2015black,doerr2018optimal}. As an implementation detail, also in line with experiments made in previous work~\cite[Section~4.1]{AntipovBD22}, we only evaluate solutions $y^{(i)}$ that are different from both, $x$ and $x'$. This implies that the number of evaluations in this crossover phase can be smaller than $\lambda_c$. 

\textbf{Selection step.} The best of all offspring, $y$ (selected again u.a.r. among all best options), replaces the parent $x$ if (and only if) it is at least as good as it.

\paragraph{Minimizing expected running time on OneMax}
We consider the DAC problem in which the objective is to minimize the expected optimization time (i.e., the expected total number of function evaluations) that the $(1+(\lambda,\lambda))$~GA requires to identify the maximum of any \onemax function $\onemax_z: \{0,1\}^{n} \to \mathbb{R}, x \mapsto |\{1 \le i \le n | x_i=z_i\}$, $z\in\{0,1\}$, a problem also referred to as \emph{Mastermind with two colors}~\cite{doerr2014playing}. This problem has spurred much interest of the runtime analysis community in parameter control since it was first empirically observed in~\cite{doerr2015black} and later formally proven in~\cite{doerr2018optimal} that an adaptation of the so-called \emph{one-fifth success rule}, a classical control policy in evolution strategies~\cite{SchumerS68,Rechenberg,Devroye72}, to discrete search spaces~\cite{kern2004learning} can achieve linear expected optimization time, a performance that is impossible to obtain with any static parameter setting~\cite{doerr2018optimal}. Follow-up works such as~\cite{AntipovBD24,AntipovBD22,BassinBS21} study different ways of selecting parameters to achieve similar speedups. 

As an implementation detail, we note that we only need to consider the optimization of the classical \onemax function $\onemax_{(1,\ldots,1)}:\{0,1\}^{n} \to \mathbb{R}, x \mapsto \sum_{i=1}^{n}x_i$ since the \onell is an unbiased algorithm in the sense of Lehre and Witt proposed in~\cite{LehreW12}. Its performance is hence identical on all functions $\onemax_z$, $z\in\{0,1\}^n$. 

\paragraph{Previous results for single-parameter control policies}

In the early works studying parameter control policies for the \onell optimizing \onemax, only $\lambda_m$ is controlled, and a static dependence of $\lambda_c =\lambda_m$ and static choice $\alpha = \beta = 1$ were used. 
%
As mentioned, a variant of the \onell equipped with the one-fifth success rule to control $\lambda_m$ was shown to yield a super-constant speed-up over any possible static choice~\cite{doerr2015black,doerr2018optimal}. In this variant, $\lambda_m$ does not depend on the quality of the current-best solution; instead it is adjusted in a success-based manner. Initialized as $\lambda_m=1$, the value of $\lambda_m$ is increased to $A \lambda_m$ when a strictly better solution is found. Here, $A >1$ is a fixed constant, referred to as the \emph{update strength}. The population size $\lambda_m$ is updated to $\beta \lambda_m$ otherwise, with $0<\beta<1$ being the second update strength. To stipulate the one-fifth success rule, one sets $A=F^{(1/4)}$ and $b=1/F$ for some fixed factor $F>1$. With this setting, $\lambda_m$ remains constant if on average one out of five iterations is successful in finding a strictly better solution. 


In an effort to automate the identification of effective parameter control policies, Chen et al.~\cite{chen2023using} considered Algorithm~\ref{alg:onell} with $\lambda_c=\lambda_m$ and $\alpha=\beta=1$ to investigate how well the automated algorithm configuration tool \textsc{irace}~\cite{lopez2016irace} would approximate a good policy for controlling $\lambda_m$. While a na\"ive application of \textsc{irace} was unable to find good policies, an iterative ``binning-and-cascading'' approach was able to learn control policies that were significantly better than all previously known policies. However, the iterative configuration approach is computationally expensive, leading Nguyen et al.~\cite{nguyen2025importance} to investigate the effectiveness of employing deep-RL, particularly DDQN, to dynamically control $\lambda_m$, resulting not only in highly effective control policies but also a significant improvement in the sampling efficiency of the learning step compared to~\cite{chen2023using}.

\paragraph{Previous results for multi-parameter control policies}

A generalization of the \onell using the one-fifth success rule was investigated in~\cite{dang2019hyper}, with the goal to understand the potential of fine-tuning the dependencies between the different parameters. As in Algorithm~\ref{alg:onell}, \cite{dang2019hyper} considers $p=\alpha \lambda_m/n$ as mutation rate and $c=\beta/\lambda_c$ as crossover bias. The crossover population size is parametrized as $\lambda_c = \gamma \lambda_m$. All in all, \cite{dang2019hyper} investigates the advantage of tuning all five parameters: $\alpha$, $\gamma$, $\beta$, $A$, and $b$ using \textsc{irace}. 
Their policy suggests $\alpha=0.3594$, $\beta=1.4128$, $\gamma=1.2379$, $A=1.1672$, and $b=0.691$. Equipped with these parameter values, the \onell was empirically shown to yield constant-factor improvements over the previously known policies, suggesting that a more flexible coupling of the five parameters can be beneficial. Note that we further extend these results by adding an additional layer of flexibility, completely decoupling the parameters from each other, and allowing each of them to be determined by the quality of the current-best solution, irrespective of the value of the other parameters.

\section{Deep-RL for Multi-parameter Control of \onell on \onemax}

\label{section:deep_rl_mp_control}
\begin{figure*}[t]
    \centering
        \includegraphics[width=0.85\linewidth, trim=0 10pt 0 10pt, clip]{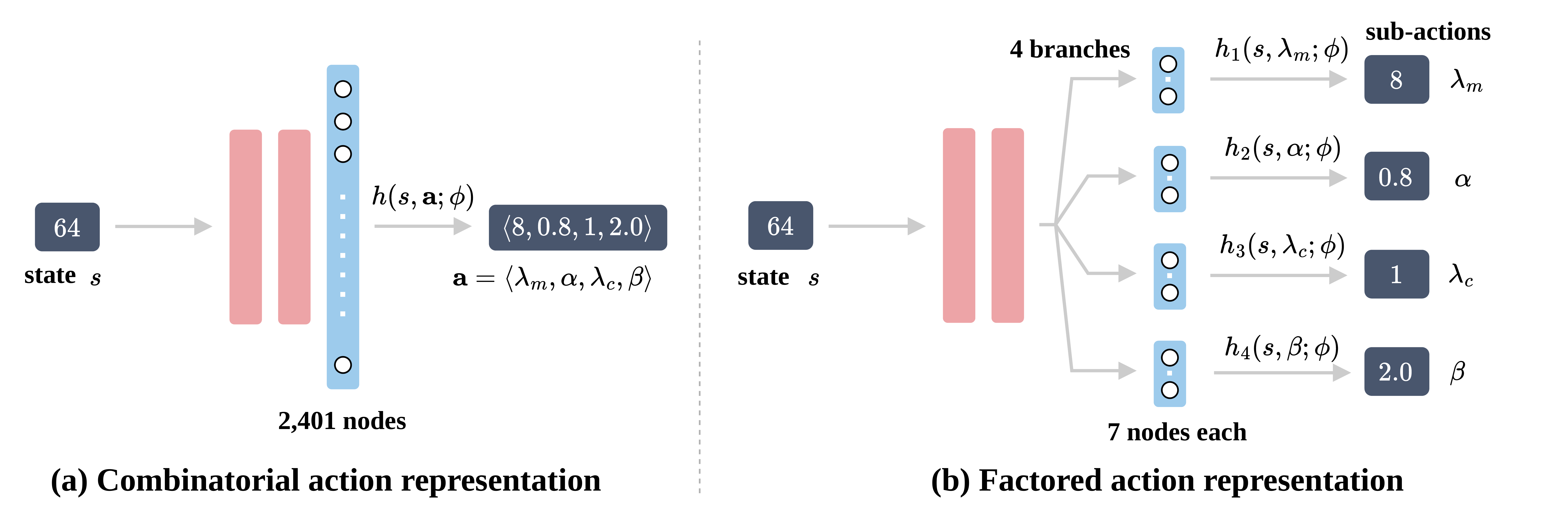}
    
    \caption{Deep neural network architectures for (a) \textbf{Combinatorial} and (b) \textbf{Factored} action space representations for the \onell with four controllable parameters, each of which is selected from a set of 7 possible values. The number of output nodes for the combinatorial representation is $7^4=2{,}401$. These diagrams depict the architectural mapping from a given state $s$ to a generic score function $h(s, \mathbf{a};\phi)$ during inference. We refer to~\Cref{subsec:comb_rep,subsec:fact_rep} for a detailed explanation of the mapping from score function $h(\cdot)$ to action $\mathbf{a}$.}
    \label{fig:mp_dac}
\end{figure*}
\textbf{Problem formulation.} When the framework of contextual MDP is extended for multi-parameter control, the function $\mathcal{M}_{\mathcal{I}}$ introduced in~\Cref{sec:backgroundRL} can be expressed as $\mathcal{M}_i := \left< \mathcal{S}, \{\mathcal{A}_d\}^{D}_{d=1}, \mathcal{T}_i, \mathcal{R}_i \right>$, with $D$-dimensional action spaces $\{\mathcal{A}_d\}^{D}_{d=1}$, and transition function $\mathcal{T}_i\colon\mathcal{S}\times\{\mathcal{A}_d\}^{D}_{d=1}\times\mathcal{S}\to[0,1]$, and reward function $\mathcal{R}_i\colon\mathcal{S}\times\{\mathcal{A}_d\}^{D}_{d=1}\to\mathbb{R}$. 

\textbf{State space.} The fitness value $f(x) \in \{0,\ldots,n-1\}$ of the current-best solution is used as the observation for deep-RL training. 

\textbf{Action space.} While PPO is naturally capable of handling both discrete and continuous action spaces with ease, DDQN requires discrete action spaces. It is therefore necessary to discretize the domains of the parameters of the \onell. Following~\cite{nguyen2025importance,nguyen2025deep}, we define the portfolio of $\lambda_m$ and $\lambda_c$ as $\mathcal{D}_1 = \{1, 2, 4, 8, 16, 32, 64 \}$. The upper bound was chosen based on results suggested in~\cite{nguyen2025importance}, where it was observed that the population size in the best policies learned by deep-RL never exceeded 64 for all tested problem sizes of $n \le 2{,}000$.\footnote{In our experiments, we observed that this choice also works for very large $n$ up to 40,000.} For $\alpha$ and $\beta$, we use $[0.25,2]$ as the domain, following the setting in~\cite{dang2019hyper}. We discretize this domain uniformly for our multi-dimensional discrete action space experiments. For simplicity, we use the same number of choices as in $\mathcal{D}_1$, i.e., $\alpha$ and $\beta$ are chosen from the set of $\mathcal{D}_2 = \{0.25, 0.542, 0.833, 1.125, 1.417, 1.708, 2\}$.

\textbf{Reward function.} We explore multiple reward functions proposed in \cite{nguyen2025importance} and further examined in~\cite{nguyen2025deep}, beginning with the conventional reward function: $r_t = -E_t + \Delta f_t$, where $E_t$ is the total number of solution evaluations in the current iteration and $\Delta f_t = f(x_{t+1}) - f(x_t)$ measures the fitness improvement. We then follow the proposed adaptive shifted variant $r_t = -E_t + \Delta f_t + b^{-}_{a}$ introduced by Nguyen et al.~\cite{nguyen2025importance}, where $b^{-}_{a}$ is an adaptive bias term computed from the replay buffer during the warm-up phase; it helps to address the under-exploration issue. In this DAC setting, the advantages of reward shifting were explicitly observed in DDQN. Its application in policy-based RL algorithms remains an open direction, although there is some evidence in the general RL literature that policy-based RL can be enhanced by reward shaping~\cite{zheng2018learning,dierkes2024combining}. In this study, we aim to address the gap in the capabilities of deep-RL algorithms in this context by examining both conventional and shifted reward functions.

In the remainder of this section, we provide a brief overview of how to engineer deep-RL algorithms to transition smoothly from single-parameter to multi-parameter control. The primary distinction lies in the action space representation, which serves as the interface between the deep-RL agent and the evolutionary algorithm. In the literature, two architectures of action space enable RL agents to interact with multiple aspects of the environment:  \emph{combinatorial} action representation and \emph{factored} action representation.

\subsection{Combinatorial Representation}
\label{subsec:comb_rep}

Combinatorial action space representation is frequently considered the simplest method because it condenses a multi-dimensional action space into a single-dimensional one by merging multiple discrete action spaces into a single list of unique combinations. 

Formulating $\left< \lambda_{m}, \alpha, \lambda_c, \beta \right>$ as a unified combinatorial action allows us to directly leverage established methods like~\cite{nguyen2025importance} to map environmental states to optimal parameter configurations. In practice, implementing this with a DDQN requires only minor architectural adjustments: the output layer of the baseline $\mathcal{Q}$-network (shown in Figure~\ref{fig:mp_dac}a) is simply widened to encompass every discrete combination in the joint action space. During inference, an agent in state $s_t$ (at step $t$) determines the best combinatorial parameter setting by computing: 

\begin{align}
\label{eq:comb_ddqn}
\mathbf{a}_t &= \argmax_{\mathbf{a}} h(s_t, \mathbf{a};\phi) \nonumber \\
 &= \argmax_{\left< \lambda_{m}, \alpha, \lambda_c, \beta \right>} \mathcal{Q}(s_t,\left< \lambda_{m}, \alpha, \lambda_c, \beta \right>; \phi)\, ,
\end{align}
where $h(\cdot)$ is the mapping function that represents the $\mathcal{Q}$-values generated by the network.

This logic extends naturally to PPO, where the actor network is structured to output a categorical probability distribution over the entire combinatorial action space using \textsc{softmax} function:
\begin{align}
\label{eq:comb_ppo_policy}
\pi(\mathbf{a} | s_t; \phi) &= \frac{\exp(h(s_t, \mathbf{a}; \phi))}{\sum_{\mathbf{a}'} \exp(h(s_t, \mathbf{a}'; \phi))} \nonumber \\
 &= \frac{\exp(h(s_t, \left\langle \lambda_{m}, \alpha, \lambda_c, \beta \right\rangle; \phi))}{\sum_{\mathbf{a}'} \exp(h(s_t, \mathbf{a}'; \phi))} \, ,
\end{align}
where the denominator sums over all valid parameter configurations $\mathbf{a}'$ in the combinatorial space. During deterministic inference, the agent selects the combined action $\mathbf{a}$ that maximize the learned policy: 

\begin{align}
\mathbf{a}_t &= \argmax_{\mathbf{a}} \pi(\mathbf{a} | s_t; \phi) \nonumber \\
\label{eq:comb_ppo}
 &= \argmax_{\left\langle \lambda_{m}, \alpha, \lambda_c, \beta \right\rangle} \pi(\left\langle \lambda_{m}, \alpha, \lambda_c, \beta \right\rangle | s_t; \phi) \, .
\end{align}


\subsection{Factored Representation}
\label{subsec:fact_rep}

Using combinatorial actions treats each unique combination of sub-actions (i.e., individual parameter values) as a distinct atomic action without considering the internal structure. 
If there are $n_d$ possible values for each sub-action and an action is composed of $D$ sub-actions $\langle a_1, \ldots, a_D \rangle$, then the total action space grows exponentially as $\prod_{d=1}^{D} n_d$. Such a combinatorial explosion poses serious challenges for RL, particularly for value-based methods like DDQN, as it hinders learning efficiency and increases inference cost~\cite{dulac2019challenges, covington2016deep, dulac1512deep, ddpg-soft-update}. To address this issue, prior work has explored hierarchical action selection~\cite{he-etal-2016-deep-reinforcement}, common-sense constraints~\cite{rasheed2020deep}, and learning-based reduction strategies~\cite{zahavy2018learn} to effectively prune the action space. To investigate the aforementioned issue, we employ a factored action structure, with assumptions of independence. 

For DDQN, similar to~\cite{tavakoli2018action}, we decompose each action into $D$ sub-actions, each of which will be modeled independently using a dedicated architecture. Specifically, for each coordinate $1 \le d \le D$, we define a distinct action-value function $\mathcal{Q}_d(s, a_d)$ that estimates the expected return of taking sub-action $a_d$ in state $s$,  based on a local $d$-th $\mathcal{Q}$-function. As a result, we can train these functions independently, each of which captures only the marginal contribution of each coordinate/parameter, without explicitly modeling interactions between sub-actions.

This decomposition dramatically reduces the effective action space size and enhances scalability. To operationalize this structure, we adopt the \emph{action branching} framework introduced by~\cite{tavakoli2018action}, which offers an efficient and empirically validated approach for learning in factored action spaces. In our implementation, we construct a shared neural network backbone (illustrated in Figure~\ref{fig:mp_dac}b) with $D = 4$ output branches (or ``\emph{heads}''), each corresponding to a generic score function $h_d(\cdot;\phi)$. At inference time, the final joint action is constructed by independently selecting the optimal value in each sub-action dimension:
\begin{align}
\label{eq:fact_ddqn}
\mathbf{a}_t &= \left\langle
\begin{array}{l}
    \argmax h_1(s_t,\lambda_{m}; \phi)\\
    \argmax h_2(s_t,\alpha; \phi)\\
    \argmax h_3(s_t,\lambda_{c};\phi)\\
    \argmax h_4(s_t,\beta;\phi)\\
\end{array}
\right\rangle\ \nonumber \\
 &= \left\langle
\begin{array}{l}
    \argmax\mathcal{Q}_1(s_t,\lambda_{m};\phi)\\
    \argmax\mathcal{Q}_2(s_t,\alpha;\phi)\\
    \argmax\mathcal{Q}_3(s_t,\lambda_{c};\phi)\\
    \argmax\mathcal{Q}_4(s_t,\beta;\phi)\\
\end{array}
\right\rangle\ \, .
\end{align}

This factored representation extends naturally to PPO, where the generic output heads $h_d(\cdot;\phi)$ represent raw log-probabilities (logits). A \textsc{softmax} activation normalizes these logits across the discrete choices for each dimension to yield independent marginal policy distributions:
\begin{equation}
\label{eq:fact_ppo_marginal}
\pi_d(a_d|s_t;\phi) = \frac{\exp(h_d(s_t, a_d;\phi))}{\sum_{a'} \exp(h_d(s_t, a';\phi))} \, .
\end{equation}

Assuming conditional independence between the sub-actions given the state $s_t$, the joint policy over the entire parameter configuration is simply the product of these marginals:
\begin{equation}
    \pi(\left< \lambda_{m}, \alpha, \lambda_c, \beta \right> | s_t; \phi) = \prod_{a_d \in \{\lambda_m, \alpha, \lambda_c, \beta \}} \pi_{d}(a_d | s_t;\phi) \, .
\end{equation}

Consequently, during deterministic inference, the aggregated action is constructed by independently selecting the mode of each marginal policy:
\begin{equation}
\label{eq:multi_dim_ppo}
\mathbf{a}_t=\left\langle
\begin{array}{l}
    \argmax\pi_{1}(\lambda_{m} | s_t;\phi)\\
    \argmax\pi_{2}(\alpha| s_t;\phi)\\
    \argmax\pi_{3}\lambda_{c}| s_t;\phi)\\
    \argmax\pi_{4}(\beta| s_t;\phi)\\
\end{array}
\right\rangle\ \, .
\end{equation}

Note that while both DDQN and PPO utilize the \textsc{argmax} operator, PPO applies it to identify the mode of the distribution, whereas DDQN uses it to determine the action with the highest expected value.

\section{Experimental Setup}

Following \cite{biedenkapp2022theory,nguyen2025importance,nguyen2025deep}, we employ a neural network function approximator with two hidden layers, each containing 50 nodes, as the feature extractor for both the DDQN and PPO algorithms. For the output heads, the DDQN with combinatorial representation consists of $7^4 = 2{,}401$ output nodes. The DDQN with a branching network and the PPO actor network share the same architecture featuring four heads, each offering seven possible choices.

We also follow previous works that employ a default setting for DDQN's hyperparameter:\footnote{The hyperparameter configuration adheres to the recommendation in~\cite{mnih2015human}, with the exception of the optimizer and learning rate, which are derived from the defaults in Stable Baselines~\cite{stable-baselines}.} an $\varepsilon$-greedy value of $0.2$, a replay buffer of $1$ million transitions, and pre-fill it with $10{,}000$ random samples. Optimization uses Adam~\cite{adam} with a batch size of $2{,}048$ and a learning rate of $0.001$. We examine the effect of the discount factor $\gamma$ on learning performance when employing DDQN algorithms. A recent work~\cite{nguyen2025deep} has highlighted the critical role of the discount factor in reward estimation, noting its sensitivity in the \onemax benchmark due to the long-horizon nature of the problem. Accordingly, high value of discount factor or undiscounted learning is advantageous in this DAC setting to prevent underestimation of future rewards. Based on these insights, we evaluate three discount factor values, $\gamma \in \{0.99, 0.9998, 1.0\}$, where the latter options apply minimal discounting to more effectively capture long-term reward signals.

For PPO, we use the Stable Baselines~\cite{stable-baselines} implementation and experiment PPO with default hyperparameters: a learning rate of $0.0003$, $2{,}048$ rollout steps, a batch size of $64$, and $10$ training epochs. We set a GAE~\cite{schulman2015high} coefficient of $0.95$, and a clip range of $0.2$, with advantage normalization for stable learning. Unlike DDQN experiments, which investigate three values of the discount factor, the PPO experiment sets the discount factor to $0.99$, as previous work~\cite{nguyen2025deep} has shown that the discount factor has less impact on PPO performance in this DAC context.

The \onemax problem can be treated as an \emph{episodic task}, in which the optimization process is naturally divided into episodes. Each episode begins from a defined initial state and ends either when the global optimum is reached (i.e., $f(x)=n$)  or when a predefined time limit is exceeded. To prevent excessive use of the training budget on ineffective policies, we impose a maximum of $0.8 n^2$ solution evaluations per \onell run, where $n$ denotes the problem size. This maximal budget is significantly larger than the linear running times we expect to see according to~\cite{doerr2018optimal} and also larger than empirical running times observed, for example, in~\cite{AntipovBD22}. Our experiments reveal that fewer than $5\%$ of RL policies result in runtimes reaching the threshold of $0.8n^2$, with most of these occurrences happening at the start of the training process when the parameters of deep-RL policies are initialized randomly.

RL training is conducted per problem size with a training budget of $200{,}000$ training time steps for each RL training run. Following previous work~\cite{nguyen2025importance} on the same benchmark, to select the best policy from each RL training, we evaluate the learned policy at every $2{,}000$ training time steps with a budget of $100$ runs per evaluation. At the end of the training, we collect the top 5 best learned policies, re-evaluate each of them $1{,}000$ times and extract the best one. As deep-RL is commonly known to be unstable~\cite{henderson2018deep,islam2017reproducibility}, and since our main focus in this work is to leverage deep-RL to discover new parameter control policies for this particular benchmark, for each RL setting, we repeat the training $5$ times and use the best policy among those trainings as the final policy. 

\textbf{Computing resources.} All experiments are conducted on a machine equipped with a single-socket AMD EPYC 7443 24-core processor. For the RL training phase we use a single core, while for the evaluation phase, we parallelize 16 threads. The total wall-time for each RL training ranges between 30 minutes to one hour, depending on the problem size. This budget is smaller than the \textsc{irace}-based approach where a tuning was reported to take up to 2 days with 20 parallel threads~\cite{dang2019hyper}.

\textbf{Performance metric.} We gather the runtimes across $1{,}000$ random seeds and report them as the Expected Runtime (ERT), which serves as our performance metric. These runtimes are defined as the total number of solution evaluations used during one \onell run. To enhance comparative analyses across problem sizes, we present the performance metric in a normalized form, dividing ERT by the problem size ($\text{ERT} / n$).

\textbf{Statistical tests.} In all result tables, the policy with the best $\text{ERT} / n$ is highlighted in \textbf{bold}. We perform a Wilcoxon signed-rank test (confidence level of $0.95$) comparing each of the remaining policies against the best one. Policies that are not statistically significantly worse than the best are \underline{underlined}. We apply the Holm-Bonferroni method to correct for multiple comparisons.

\textbf{Baselines.} We compare our deep-RL multi-parameter control policies against several baselines from the literature as outlined in~\Cref{sec:backgroundga}, including:
\begin{itemize}
    \item the theory-derived single-parameter control policy proposed in~\cite{doerr2015black,doerr2018optimal}, namely $\pi_{\textsc{theory}}$, where $\lambda_m=\sqrt{n/(n-f(x))}$.
    \item the single-parameter control policy based on the one-fifth success rule~\cite{kern2004learning,auger2009benchmarking};
    \item the multi-parameter control policy obtained from \textsc{irace}~\cite{dang2019hyper};
    \item the two single-parameter control policies based on the two deep-RL methods, $\pi_{\text{DDQN/sp}}$ and $\pi_{\text{PPO/sp}}$, obtained in~\cite{nguyen2025importance,nguyen2025deep}.\footnote{The ERT reported in~\cite{nguyen2025importance} was based on average performance across five RL training runs.  However, in our work, we report the best policy because we will use it to derive a symbolic policy. To ensure a fair comparison, we re-run their experiments and report performance of the best policy across five RL training runs instead of taking the average.}
\end{itemize}

\begin{table}[t]
    \centering
\caption{Normalized ERT (and its standard deviation) of the best policies (DDQN only) achieved by each combination of  \textcolor{combinatorialcolor}{\textbf{C}}ombinatorial and \textcolor{factoredcolor}{\textbf{F}}actored action representations (AR), Na\"ive and Adaptive Shifting (AS) reward functions, and discount factors ($\gamma$).  \textbf{Bold}: the best normalized ERT. \underline{Underlined}: not significantly worse than the best (confidence level of $0.95$).}
    \label{tab:compare_rl_reward_gamma}
    \begin{tabular}{llcccc}
        \toprule
        & & & $n=\mathbf{100}$ & $n=\mathbf{200}$ & $n=\mathbf{500}$ \\
        \midrule
        \multicolumn{3}{l}{$\pi_\textsc{theory}$~\cite{doerr2015black,doerr2018optimal}}&5.826\scriptsize{(1.18)}&6.167\scriptsize{(0.97)}&6.474\scriptsize{(0.67)}\\

        \midrule
        \rowcolor{gray!20} \emph{AR} & \emph{Reward} & $\gamma$ & & & \\
        \midrule
\textcolor{combinatorialcolor}{\textbf{C}}&Na\"ive&$0.99$&4.510\scriptsize{(1.07)}&5.057\scriptsize{(0.83)}&5.959\scriptsize{(0.79)}\\
\textcolor{combinatorialcolor}{\textbf{C}}&AS&$0.99$&4.392\scriptsize{(0.77)}&4.961\scriptsize{(0.76)}&5.583\scriptsize{(0.74)}\\
\textcolor{combinatorialcolor}{\textbf{C}}&Na\"ive&$0.9998$&4.350\scriptsize{(0.80)}&4.969\scriptsize{(0.72)}&6.531\scriptsize{(0.61)} \\
\textcolor{combinatorialcolor}{\textbf{C}}&AS&$0.9998$&4.777\scriptsize{(0.81)}&4.835\scriptsize{(0.59)}&5.198\scriptsize{(0.51)}\\

\textcolor{combinatorialcolor}{\textbf{C}}&Na\"ive&$1.0$&4.528\scriptsize{(0.82)}&4.712\scriptsize{(0.69)}&5.369\scriptsize{(0.74)}\\

\textcolor{combinatorialcolor}{\textbf{C}}&AS&$1.0$&4.584\scriptsize{(0.80)}&5.326\scriptsize{(0.60)}&5.843\scriptsize{(0.55)}\\

\textcolor{factoredcolor}{\textbf{F}}&Na\"ive&$0.99$&4.952\scriptsize{(1.38)}&5.731\scriptsize{(1.42)}&6.835\scriptsize{(1.47)}\\
\textcolor{factoredcolor}{\textbf{F}}&AS&$0.99$&4.395\scriptsize{(0.97)}&4.689\scriptsize{(0.79)}&5.948\scriptsize{(0.93)}\\
\textcolor{factoredcolor}{\textbf{F}}&Na\"ive&$0.9998$&4.467\scriptsize{(0.98)}&\underline{4.472}\scriptsize{(0.63)}&4.932\scriptsize{(0.52)}\\
\textcolor{factoredcolor}{\textbf{F}}&AS&$0.9998$&4.318\scriptsize{(0.83)}&\underline{4.483}\scriptsize{(0.65)}&\textbf{4.814}\scriptsize{(0.42)}\\

\textcolor{factoredcolor}{\textbf{F}}&Na\"ive&$1.0$&4.518\scriptsize{(0.99)}&\textbf{4.449}\scriptsize{(0.71)}&4.848\scriptsize{(0.48)}\\
\textcolor{factoredcolor}{\textbf{F}}&AS&$1.0$&\textbf{4.189}\scriptsize{(0.81)}&4.564\scriptsize{(0.74)}&4.938\scriptsize{(0.48)}\\

        \bottomrule
    \end{tabular}%
\end{table}%

Note that we do not explicitly compare to the results presented in~\cite{AntipovBD22} and~\cite{AntipovBD24} since both are much worse than those obtained by $\pi_{\textsc{theory}}$, with ERTs exceeding $12n$ for~\cite{AntipovBD22} and results worse than the (1+1)~EA (and hence much worse than the \onell equipped with $\pi_{\textsc{theory}}$) in~\cite{AntipovBD24}. Among the baselines, the \textsc{irace}-based approach~\cite{dang2019hyper} is the strongest one. This is not too surprising, as this is the only baseline that tunes all four parameters of \onell separately. 

\section{Results}
\label{sec:ddqn_for_onemaxmpdac}

\textbf{Comparison of DDQN Approaches.} 
 \label{subsec:comparison_ddqn}
We first evaluate the impact of the three DDQN's design choices, including: (i)  action space representation (combinatorial vs. factored); (ii) reward function (na\"ive vs. adaptive reward shifting, as in~\cite{nguyen2025importance}); and (iii) various values of discount factor.

\Cref{tab:compare_rl_reward_gamma} presents the performance of all combinations of those three design alternatives on three problem sizes $n \in \{100; 200; 500\}$. Interestingly, when the na\"ive reward function and the commonly used discount factor $\gamma=0.99$ are used, there is no clear advantage in using factored action space representation. In contrast, when adaptive reward shifting and/or higher discount factor is adopted, the factored representation starts outperforming its combinatorial counterpart across all problem sizes (as shown in the last four rows of \Cref{tab:compare_rl_reward_gamma}). Additionally, the learned policies are statistically significantly superior to all others.

These observations significantly enhance the findings in~\cite{nguyen2025deep}, where the higher discount factor (e.g., $\gamma=0.9998$ or $1.0$) was considered only as an alternative to the adaptive reward shifting mechanism in the long-horizon environment. Their combination has previously been seen as an implicit conflict, resulting in performance degradation. However, the results in this section offer a new perspective, demonstrating that these two enhancements no longer conflict but can actually support each other in this multi-parameter control benchmark.

\textbf{Policy Collapse in PPO.} 
\label{subsec:policy_collapse_ppo}
While the advantages of reward shaping in policy-gradient-based RL have been discussed in the literature, particularly in general RL applications (i.e., see~\cite{zheng2018learning}), this remains an open question in this DAC context, across both single- and multi-dimensional action spaces. In this section, we examine the performance of PPO across three axes: i) single- versus multi-parameter controls; ii) combinatorial versus factored action space representations in the case of multi-parameter control; and iii) conventional versus shifted reward functions. However, unlike the reward shifting successfully applied in DDQN, the adaptive shifting mechanism proposed in~\cite{nguyen2025importance} is tailored specifically for DDQN as it relies on the replay buffer of the off-policy RL family. Although we cannot elegantly apply adaptive reward shifting in PPO, we still conduct it through a brute-force search by testing a grid of potential negative shifting bias\footnote{As shown in~\cite{nguyen2025importance}, adding a negative shifting bias into the conventional reward function can induce exploration.} values such as $b\in\{0,-1,-3,-5,-7\}$, where $b=0$ denotes the conventional reward function.

\begin{table}[t]
    \centering
\caption{Normalized ERT (and its standard deviation) of the best policies (PPO only) across parameter control (Single versus Multiple), action space representation (AR), including \textcolor{combinatorialcolor}{\textbf{C}}ombinatorial versus \textcolor{factoredcolor}{\textbf{F}}actored, in multi-parameter control, and reward shifting bias. \textbf{Bold}: the best normalized ERT. \underline{Underlined}: not significantly worse than the best (confidence level of $0.95$).}
    \label{tab:compare_ppo_reward_shifting}
    \begin{adjustbox}{max width=0.9\textwidth} 
    \begin{tabular}{lcrccc}
        \toprule
         & & & $n=\mathbf{100}$ & $n=\mathbf{200}$ & $n=\mathbf{500}$ \\
        \midrule 
        \multicolumn{3}{l}{$\pi_\textsc{theory}$~\cite{doerr2015black,doerr2018optimal}}&5.826\scriptsize{(1.18)}&6.167\scriptsize{(0.97)}&\textbf{6.474}\scriptsize{(0.67)}\\
        \midrule
        \rowcolor{gray!20} \emph{Parameter} & \emph{AR} & \emph{Bias} & & & \\
        \midrule
        \multirow{5}{*}{Single} & &$0$&6.147\scriptsize{(1.61)}&7.003\scriptsize{(1.65)}&7.572\scriptsize{(0.97)}\\
        &&$-1$&6.183\scriptsize{(1.71)}&6.642\scriptsize{(1.07)}&8.009\scriptsize{(0.95)}\\
        &&$-3$&6.138\scriptsize{(1.66)}&6.568\scriptsize{(1.10)}&7.031\scriptsize{(0.99)}\\
        &&$-5$&6.019\scriptsize{(1.27)}&5.967\scriptsize{(1.14)}&8.333\scriptsize{(1.61)}\\
        &&$-7$&6.078\scriptsize{(1.13)}&6.020\scriptsize{(1.08)}&8.142\scriptsize{(1.57)}\\
        \arrayrulecolor{lightgray}\midrule

        \multirow{5}{*}{Multiple} &  \textcolor{combinatorialcolor}{\textbf{C}} &$0$&\underline{4.962}\scriptsize{(1.45)}&5.791\scriptsize{(1.44)}&6.642\scriptsize{(1.20)}\\
        &\textcolor{combinatorialcolor}{\textbf{C}}&$-1$&5.327\scriptsize{(1.53)}&5.840\scriptsize{(1.45)}&6.847\scriptsize{(1.51)}\\
        &\textcolor{combinatorialcolor}{\textbf{C}}&$-3$&\underline{4.957}\scriptsize{(1.41)}&\underline{5.755}\scriptsize{(1.39)}&6.876\scriptsize{(1.40)}\\
        &\textcolor{combinatorialcolor}{\textbf{C}}&$-5$&\underline{4.968}\scriptsize{(1.45)}&6.147\scriptsize{(1.41)}&6.694\scriptsize{(1.38)}\\
        &\textcolor{combinatorialcolor}{\textbf{C}}&$-7$&\underline{4.908}\scriptsize{(1.43)}&\underline{5.737}\scriptsize{(1.47)}&6.718\scriptsize{(1.37)}\\
        \arrayrulecolor{lightgray}\midrule
        \multirow{5}{*}{Multiple} &  \textcolor{factoredcolor}{\textbf{F}} &$0$&\underline{4.905}\scriptsize{(1.32)}&\underline{5.691}\scriptsize{(1.45)}&6.704\scriptsize{(1.33)}\\
        &\textcolor{factoredcolor}{\textbf{F}}&$-1$&\underline{4.878}\scriptsize{(1.42)}&\underline{5.652}\scriptsize{(1.41)}&6.715\scriptsize{(1.36)}\\
        &\textcolor{factoredcolor}{\textbf{F}}&$-3$&\underline{4.906}\scriptsize{(1.40)}&\underline{5.692}\scriptsize{(1.37)}&6.716\scriptsize{(1.43)}\\
        &\textcolor{factoredcolor}{\textbf{F}}&$-5$&\underline{4.910}\scriptsize{(1.39)}&\textbf{5.649}\scriptsize{(1.33)}&6.707\scriptsize{(1.44)}\\
        &\textcolor{factoredcolor}{\textbf{F}}&$-7$&\textbf{4.863}\scriptsize{(1.39)}&\underline{5.668}\scriptsize{(1.34)}&6.692\scriptsize{(1.39)}\\
        \arrayrulecolor{black}\bottomrule
    \end{tabular}%
    \end{adjustbox}%
\end{table}%

We present performances of PPO across three problem sizes: $\{100,200,500\}$ in~\Cref{tab:compare_ppo_reward_shifting}. Interestingly, all multi-parameter control policies, whether enhanced by reward shifting or not, and regardless of the action space representation, outperform the theory-derived policy in the first two problem sizes. However, no single-parameter control policies, even with reward shifting, achieve this level of advantage. Although the dominance of multi-parameter control policies cannot be maintained for a problem size of $n=500$, these are still capable of discovering better policies (ERTs are closer to the theory-derived policy) than any single-parameter control policies. Regarding the impact of reward shifting biases, they almost have no effect in both single- and multi-parameter settings. Concretely, in the first two problem sizes in multi-parameter control using factored action space representation, where all shifting biases yield identical results in the statistical test. 

\begin{figure}[t]
  \centering
  
    \includegraphics[width=0.48\columnwidth]{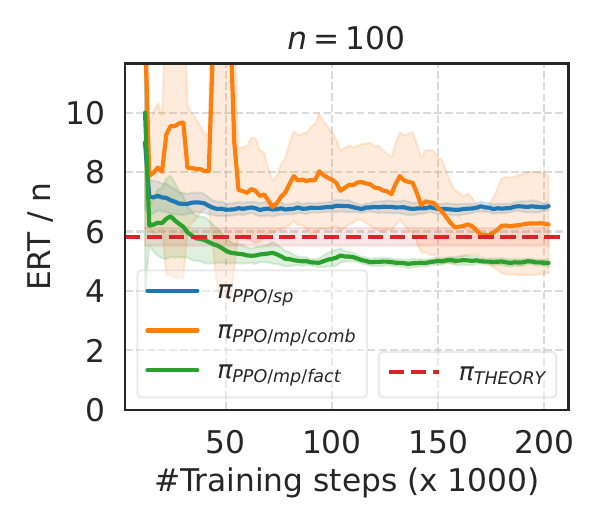}%
    \label{fig:sub1}%
    \includegraphics[width=0.48\columnwidth]{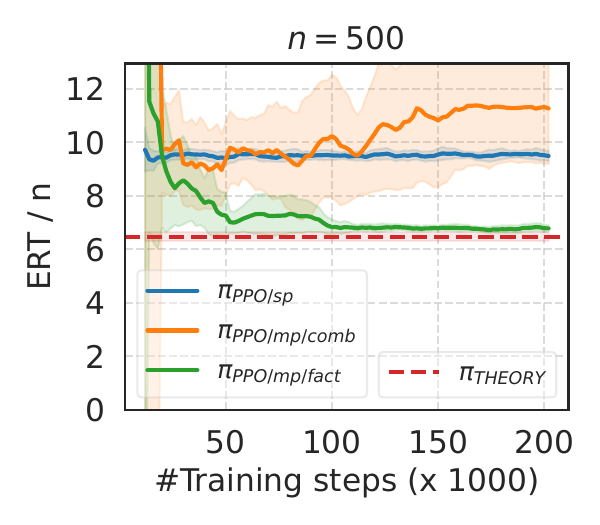}%
    \label{fig:sub2}%
  
  \caption{Learning curves for PPO under single-parameter control (\textcolor{ppo_sp}{$\pi_{\text{PPO/sp}}$}) and multi-parameter control with combinatorial (\textcolor{ppo_mp}{$\pi_{\text{PPO/mp/comb}}$}) factored (\textcolor{ddqn_mp}{$\pi_{\text{PPO/mp/fact}}$}) action space representations for two problem sizes. The solid lines represent the mean normalized ERTs, while the shaded areas indicate the standard deviation across 5 PPO runs.}
  \label{fig:sp_mp_ppo_n100_500}
\end{figure}

Although the shifted reward function does not offer any benefits for PPO, the impact of multi-parameter control is quite evident. To investigate whether this improvement is due to mere chance or genuine effectiveness, we plot the PPO learning curves for two problem sizes of $\{100; 500\}$, where $n=100$ demonstrates the predominance of the multi-parameter control policy, while $n=500$ exhibits performance fairly competitive with the theory-derived policy. It should be noted that the normalized ERTs in~\Cref{tab:compare_ppo_reward_shifting} are derived from extensive testing (over $1{,}000$ runs) of the best policy found during training, whereas each recorded point on the learning curve, shown in~\Cref{fig:sp_mp_ppo_n100_500}, represents the mean normalized ERT (over 100 runs) during training across 5 random seeds. This implies that thorough testing will be carried out on the cost-effective, less accurate policy identified on the learning curve, resulting in some differences between them. Remarkably, the learning curves of the combinatorial action space representation are significantly less stable than the factored representation. In $n=100$, although some of running seeds in the combinatorial representation setting appear to discover a good policy more quickly than the factored one within the first $50,000$ training steps, as indicated by the shaded orange area, they subsequently experience a rapid collapse and exhibit unstable learning until the end. In a more challenging problem size of $n=500$, this collapse is evident; those PPO policies stagnate at a suboptimal policy by the end of the training process (starting from training step $150,000$). In contrast, the factored action space representation consistently demonstrates learning stability across both problem sizes. We conclude that, although the factored representation cannot guarantee that PPO consistently outperforms the theory-derived policy, it can help stabilize the training process to some extent. 

\begin{figure}[t]
    \centering
        \includegraphics[width=1.0\linewidth, clip]{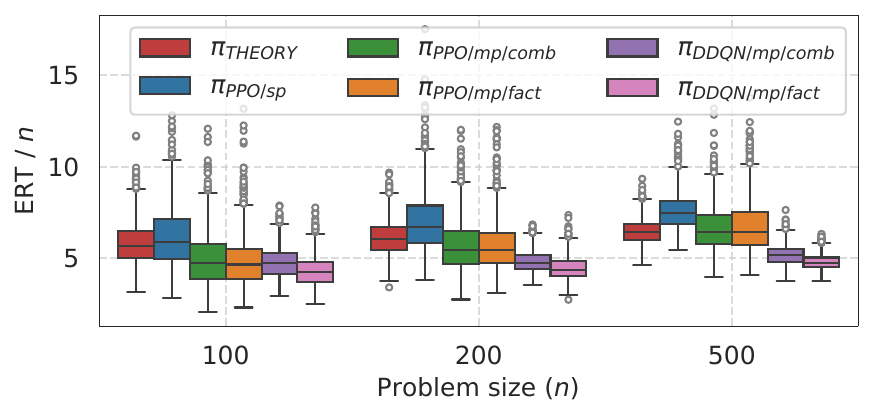}
    \caption{Normalized ERT ($\downarrow$) comparison of deep-RL policies against the theory-derived baseline $\pi_\textsc{theory}$ for problem sizes $n \in \{100, 200, 500\}$. We evaluate single-parameter PPO \cite{nguyen2025deep}, our multi-parameter PPO variants, and the top DDQN policies from \Cref{tab:compare_rl_reward_gamma}.}
    \label{fig:sp_mp_ppo_ert_boxplot}
\end{figure}

Nonetheless, upon closer examination of the learned PPO policies in both action space representations, we find that neither converges to a meaningful policy. Once the learning curves stabilize, we observe that the best learned PPO policy is $\lambda_m = 1$, $\alpha = 0.25$, $\lambda_c = 1$, and these parameters are kept constant across all fitness states, while $\beta$ tends to fluctuate randomly with values greater than $1$ (which is permissible, as the crossover bias $\beta / \lambda_c$ is bounded within $1$). This observation highlights a policy collapse in PPO method, indicating that PPO becomes fixated on an ineffective policy instead of genuinely converging to a more performant one. This finding aligns with the investigation in~\cite{nguyen2025deep}, where the PPO policies persistently sample the population size of $\lambda_m = 1$. Despite attempting various RL enhancement techniques to address this challenge, such as entropy regularization~\cite{haarnoja2017reinforcement,ahmed2019understanding}, reward normalization~\cite{engstrom2019implementation}, undiscounted setting~\cite{li2022challenges}, and even employing Hyperparameter Optimization (HPO) for tuning seven crucial hyperparameters, none of these methods appear to be effective. Therefore, we acknowledge that the policy collapse issue of PPO inherently occurs in both single- and multi-parameter settings.

Finally, when comparing the two most commonly used RL methods side-by-side, despite the improvements in PPO with factored action space representation in the multi-parameter setting, it still falls short of DDQN approaches. As shown in~\Cref{fig:sp_mp_ppo_ert_boxplot}, we visualize the normalized ERTs of all the RL policies we have evaluated. Across all three problem sizes tested, DDQNs consistently emerge as the dominant approach, especially when equipped with the three components discussed in the previous section.

\begin{table}[t]
    \centering
    \caption{Normalized ERT (and its standard deviation) of several baselines from the literature and the best policies achieved through deep-RL methods in single- and multi-parameter control (with \underline{fact}ored action space representation), using three types of reward functions: na\"ive, \underline{a}daptive, and \underline{f}ixed \underline{s}hifting. \textbf{Bold}: the best normalized ERT.}
    \label{tab:compare_all_methods}
    \begin{adjustbox}{max width=0.49\textwidth} 
    \begin{tabular}{lllccc}
        \toprule
        & & & $n=\mathbf{1{,}000}$ & $n=\mathbf{1{,}500}$ & $n=\mathbf{2{,}000}$ \\
        \midrule
        \multicolumn{3}{l}{$\pi_\textsc{theory}$~\cite{doerr2015black,doerr2018optimal}}& 6.587\scriptsize{(0.53)} & 6.647\scriptsize{(0.44)} & 6.681\scriptsize{(0.39)} \\
        \multicolumn{3}{l}{\irace~\cite{dang2019hyper}} & 5.587\scriptsize{(0.35)} & 5.621\scriptsize{(0.29)} & 5.666\scriptsize{(0.26)} \\
        \multicolumn{3}{l}{\textsc{one-fifth}~\cite{kern2004learning,auger2009benchmarking}} & 6.886\scriptsize{(0.54)} & 6.931\scriptsize{(0.48)} & 7.008\scriptsize{(0.41)} \\
        \midrule
        \rowcolor{gray!20} \emph{Deep-RL} & \emph{Parameter} & \emph{Reward} & & & \\
        \midrule
        
        DDQN & Single & AS & 6.338\scriptsize{(0.55)} & 6.209\scriptsize{(0.41)} & 6.608\scriptsize{(0.41)} \\
        PPO & Multi/Fact & Na\"ive & 7.544\scriptsize{(1.39)} & 7.938\scriptsize{(1.50)} & 8.294\scriptsize{(1.44)} \\
        PPO & Multi/Fact & FS & 7.016\scriptsize{(0.60)} & 7.940\scriptsize{(1.45)} & 8.267\scriptsize{(1.42)} \\
        \rowcolor{cyan!10} DDQN & Multi/Fact & AS & \textbf{5.397}\scriptsize{(0.41)} & \textbf{4.971}\scriptsize{(0.29)} & \textbf{5.162}\scriptsize{(0.29)} \\
        \bottomrule
    \end{tabular}%
    \end{adjustbox}%
\end{table}

\textbf{Extensive Comparative Analysis.} 
Our primary objective is to identify the best-fit deep-RL algorithm for this DAC environment as the foundation for policy discovery later. Above, we have gained initial insights with small-scale instances: (i) the combination of factored representation, adaptive reward shifting, and a higher discount factor value of $\gamma = 0.9998$ performs as the best overall configuration for DDQN~(\Cref{subsec:comparison_ddqn}); and (ii) while the combination of factored action space representation shows promising results, the learned PPO policies consistently collapse to a suboptimal policy~(\Cref{subsec:policy_collapse_ppo}). In this section, we extend the experiments using the best settings for each RL algorithm obtained in the previous sections on a broader range of problem sizes, with $n \in \{1{,}000; 1{,}500; 2{,}000\}$, shown in~\Cref{tab:compare_all_methods}.
    
Although reward shifting has not shown any impact in PPO experiments, we aim to confirm these findings on a larger scale. Therefore, we choose the combination of factored representation and two reward functions, one conventional and another with a fixed bias shift of $b=-7$, for our extensive analysis. Unexpectedly, the shifted reward function demonstrates an improvement over the conventional reward for a problem size of $n=1{,}000$; however, they quickly become identical as the problem size increases in~\Cref{tab:compare_all_methods}. It is not surprising that PPO policies consistently fail to explore any useful policies compared to the simplest baseline of theory-derived policy. 


Unlike the collapse of PPO, DDQN equipped with factored action space representation, adaptive reward shifting, and a higher discount factor (as discussed in~\Cref{subsec:comparison_ddqn}) consistently demonstrate its advantage in large-scale problem sizes, highlighted in blue in~\Cref{tab:compare_all_methods}. This setting even outperforms the strongest baseline using \textsc{irace} for tuning the same parameters. This finding aligns with the conclusion in~\cite{nguyen2025deep}, where DDQN is suggested as the best practice for RL-based DAC in this benchmark. The extensive analysis further reinforces the established insights and effectively supports the symbolic policy discovery process.

\section{Ablation Study on Parameter Importance}
\label{sec:importance_ablation_study}

\begin{table}[t]
    \centering
\caption{Normalized ERT (and its standard deviation) of different versions with varying numbers of controlled parameters. The term $\mathcal{Q}_d$ denotes dynamic control achieved through DDQN (see~\Cref{eq:fact_ddqn}), whereas $\alpha=\beta=1$ and $\lambda_c=\lambda_m$ indicate no dynamic control. \textbf{Bold}: the best normalized ERT. \underline{Underlined}: not significantly worse than the best (confidence level of $0.95$).}
    \label{tab:ablation_important_params}
    \begin{adjustbox}{max width=0.495\textwidth} 
    \begin{tabular}{cccccccc}
        \toprule
        &&&& $n=\mathbf{500}$ & $n=\mathbf{1{,}000}$ & $n=\mathbf{1{,}500}$ & $n=\mathbf{2{,}000}$ \\
        \midrule 
        \multicolumn{4}{l}{$\pi_{\textsc{theory}}$~\cite{doerr2015black,doerr2018optimal}}& 6.474\scriptsize{(0.67)}&6.587\scriptsize{(0.53)}&6.647\scriptsize{(0.44)}&6.681\scriptsize{(0.39)} \\
        \midrule
        \rowcolor{gray!20} $\lambda_m$ & $\alpha$ & $\lambda_c$ & $\beta$ & & & & \\
        \midrule
        $\mathcal{Q}_1$ & 1 & $\lambda_m$ & 1 & 6.017\scriptsize{(0.63)}&6.338\scriptsize{(0.55)}&6.209\scriptsize{(0.41)}&6.608\scriptsize{(0.41)} \\
        \arrayrulecolor{lightgray}\midrule
        $\mathcal{Q}_1$& 1 & $\mathcal{Q}_3$ & 1 &5.052\scriptsize{(0.48)}&5.268\scriptsize{(0.38)}&5.504\scriptsize{(0.39)}&6.133\scriptsize{(0.39)}\\
        $\mathcal{Q}_1$& $\mathcal{Q}_2$ & $\lambda_m$ & 1 &5.064\scriptsize{(0.60)}&5.187\scriptsize{(0.37)}&5.282\scriptsize{(0.39)}&5.277\scriptsize{(0.34)}\\
        $\mathcal{Q}_1$& 1 & $\lambda_m$ & $\mathcal{Q}_4$ &6.170\scriptsize{(0.67)}&6.220\scriptsize{(0.53)}&6.384\scriptsize{(0.44)}&6.722\scriptsize{(0.44)}\\
        \midrule
        $\mathcal{Q}_1$& $\mathcal{Q}_2$ & $\lambda_m$ & $\mathcal{Q}_4$&5.191\scriptsize{(0.63)}&5.303\scriptsize{(0.43)}&5.609\scriptsize{(0.33)}&5.410\scriptsize{(0.32)}\\
        $\mathcal{Q}_1$& $\mathcal{Q}_2$ & $\mathcal{Q}_3$ & 1 &4.738\scriptsize{(0.40)}&\textbf{4.869}\scriptsize{(0.34)}&5.196\scriptsize{(0.34)}&5.297\scriptsize{(0.32)}\\
        \midrule
         $\mathcal{Q}_1$ &  $\mathcal{Q}_2$ & $\mathcal{Q}_3$ &  $\mathcal{Q}_4$&\textbf{4.814}\scriptsize{(0.42)}&\underline{5.397}\scriptsize{(0.41)}&\textbf{4.971}\scriptsize{(0.29)}&\textbf{5.162}\scriptsize{(0.29)}\\

        \arrayrulecolor{black}\bottomrule
    \end{tabular}%
    \end{adjustbox}%
\end{table}%

We analyze the significance of each parameter to decipher the policies learned by the DDQN and provide the interpretable insights for the symbolic policy discovery. This investigation is conducted on four problem sizes: $n \in \{500; 1{,}000; 1{,}500; 2{,}000\}$.

\begin{figure}[t]
    \centering
        \includegraphics[width=0.8\linewidth, clip]{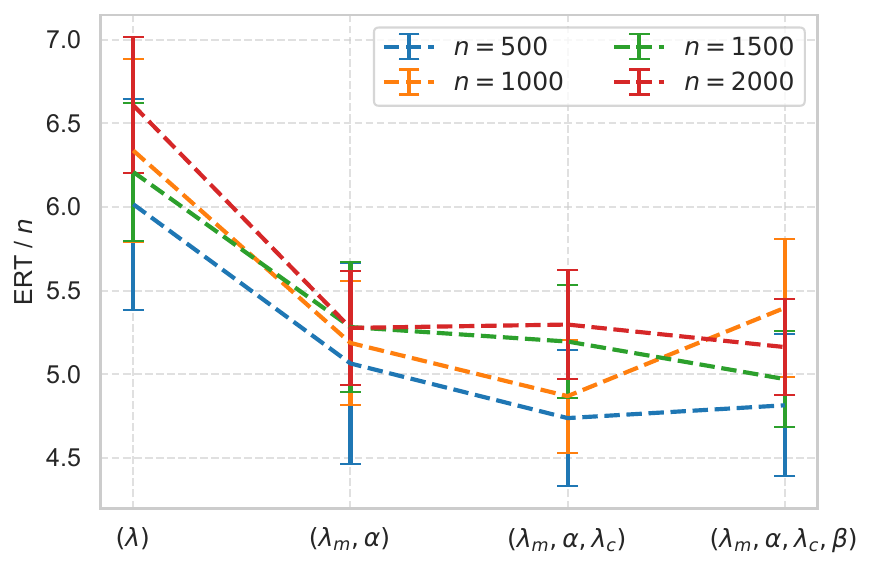}
        
    \caption{Transition from controlling only one parameter of \onell on \onemax to controlling all four parameters with DDQN across different problem sizes. Each point shows the normalized ERT (and its standard deviation) of the corresponding learned policy.}
    \label{fig:ablation_replacement_retraining}
    \vspace{-6pt}
\end{figure}

We repeat the DDQN training to control only the mutation population size ($\lambda_m$), while the remaining parameters $\{\alpha,\lambda_c,\beta\}$ are determined by the theory-derived policy $\pi_{\textsc{theory}}$, i.e., $\lambda_c=\lambda_m$ and $\alpha=\beta=1$. Gradually, we integrate more parameters into the set of controllable parameters by DDQN. For instance, in the second round, we conduct DDQN trainings for three possible pairs: $\{\lambda_m$,$\lambda_c\}$, $\{\lambda_m$,$\alpha\}$, and $\{\lambda_m$,$\beta\}$. We then select the most effective combination as the starting point for the third step, and so on. This is repeated until all four parameters are included. Table~\ref{tab:ablation_important_params} shows the performance of all combinations in each round across the four problem sizes, and a summary version, displaying only the best combination at each iteration, is presented in~\Cref{fig:ablation_replacement_retraining}.

When comparing the theory-derived policy and the policies obtained from simultaneously controlling mutation population size and the mutation rate coefficient, the average improvements achieved are over $18\%$. Although the average expected runtimes of $\{\lambda_m,\alpha\}$ do not always outperform $\{\lambda_m,\lambda_c\}$, our Wilcoxon signed-rank test's results suggest that there is no statistical difference between these two combinations. This observation highlights the crucial role of $\alpha$ in almost four tested problem sizes, and we conclude that $\alpha$ is the next parameter that should be controlled. Therefore, we then verify the impact of integrating one of the two parameters of the crossover phase into the combination. The advancements increase over $22\%$ when we incorporate the offspring population size of crossover. On the contrary, replacing $\lambda_c$ with $\beta$ adversely affects the overall performance. Controlling the pair of $\{\lambda_m,\beta\}$ also leads to the same observations, suggesting that dynamically adjusting the factor of crossover bias is not an effective way to enhance the optimization process. 

The common trend across four problem sizes is shown in~\Cref{fig:ablation_replacement_retraining}, where we observe an overall improvement in most cases when a new parameter is added into the set of controllable parameters. There are only a few exceptions, such as the last point in the transition path of $n=1{,}000$ where the performance got worse. We attribute those cases to DDQN's learning instability and the challenges when the size of the action space is increased. This observation is not too surprising as this kind of learning instability behavior has been observed in previous work on theory-derived dynamic algorithm configuration benchmarks~\cite{biedenkapp2022theory,nguyen2025importance}. We leave for future work a more thorough investigation into improving the learning stability in such cases. Overall, we observe intense drops during the transition from controlling $\lambda_m$ alone to $\{\lambda_m,\alpha\}$, followed by $\{\lambda_m,\alpha,\lambda_c\}$, and finally, we achieve full control over four parameters. This order reflects the important degree of each parameter: $\lambda_m \gg \alpha \gg \lambda_c \gg \beta$. 

\begin{figure}[t]
    \centering
    \includegraphics[width=\linewidth, clip]{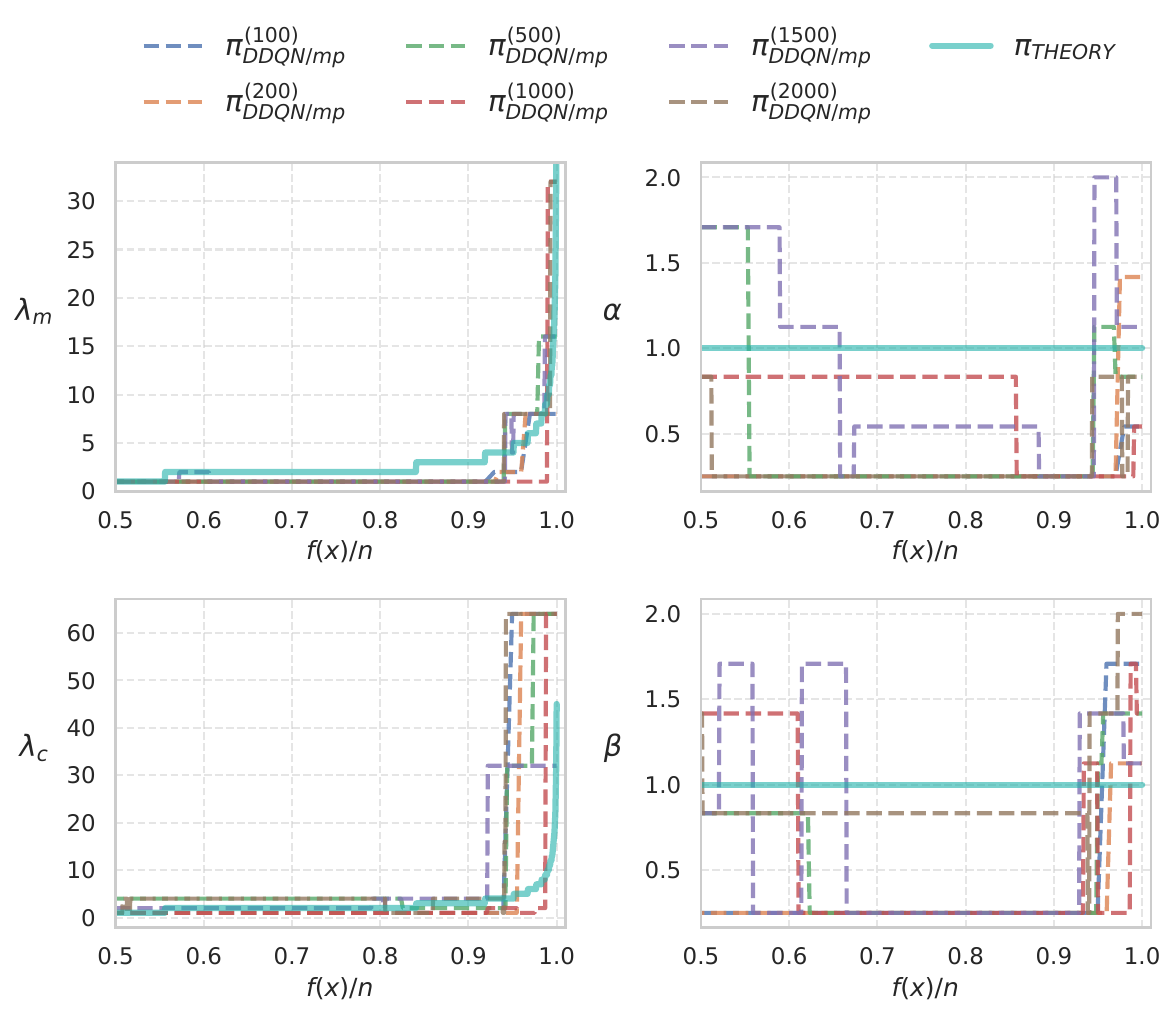}
    \caption{DDQN-based policies and the theory-derived policy across six problem sizes. Note that the evolution path of $\lambda_m$ and $\lambda_c$ in the theory-derived policy $\pi_{\textsc{theory}}$ is exactly the same for all problem sizes, however, the upper bound for those parameters increases as problem size increases. Here we plot those values to the maximum upper bound, which corresponds to $n=2{,}000$. 
    }
    \label{fig:merged_policies}
\end{figure}

\section{From Black-box to White-box Control Policies}
\label{sec:black_to_white_box}

Although the policies learned via RL show strong performance, they are problem-dependent. Moreover, as a neural network-based approach, each learned policy functions is a black box function. Those aspects may hinder the usefulness of our findings for theoretical insights. To address this limitation, in this section, based on the parameter importance analysis and the policies suggested by DDQN, we derive a new symbolic policy that is not only interpretable but also exhibit strong performance and generalization. This symbolic policy is derived through a two-stage process. We first manually design a symbolic parameter control policy based on the empirical data of the policies learned by DDQN. This policy is then further refined via a fine-tuning process using SMAC3~\cite{lindauer2022smac3}. As demonstrated later in this section, both the hand-crafted and the tuned policies consistently outperform all existing baselines, even on very large problem sizes of up to $n=40{,}000$.


\subsection{Hand-Crafted Design for Multi-Parameter Control}
\label{sec:derived_mp_policy}

We begin the process by designing a hand-crafted policy using policies learned by DDQN for all problem sizes tested in the previous section. We plot all the policies on the same canvas by normalizing fitness values and then compare them with the theory-derived policy, as shown in \Cref{fig:merged_policies}. The design process prioritizes the parameters based on their significance, as explored in~\Cref{sec:importance_ablation_study}, with the more valuable parameters being designed first.

\textbf{Population size $\lambda_m$ for the mutation phase}. Overall, this parameter was set as one by DDQN for most of the fitness range (as shown in~\Cref{fig:merged_policies}, top-left). Only later in the search, where the fitness value is around $0.95n$ on average, that $\lambda_m$ starts increasing. The magnitudes of $\lambda_m$ in that later part are fairly close to the values suggested by theory (i.e., $\lambda_m=\sqrt{n/(n-f(x))}$). The symbolic representation of this parameter is as follows:

\begin{equation}
    \label{eq:handcrafted_lbdm}
    \lambda_m =
    \begin{cases}
    1 & \text{if } f(x)/n \le 0.95 \\
    \sqrt{n/(n-f(x))}   & \text{otherwise}
    \end{cases}
\end{equation}

\begin{table*}[ht]
    \centering
\caption{Normalized ERT (and its standard deviation) for seven problem sizes of different versions of the newly hand-crafted policy. Each term in Eq.$(\cdot)$ corresponds to the equation of the new symbolic form for each parameter. The cases where $\alpha=\beta=1$ and $\lambda_c=\lambda_m$ denote no dynamic control. \textbf{Bold}: the best normalized ERT.}
    \label{tab:mpdac_derived_policies}
    \begin{adjustbox}{max width=\textwidth} 
    \begin{tabular}{ccccccccccc}
        \toprule
        &&&& $n=\mathbf{3{,}000}$ & $n=\mathbf{4{,}000}$ & $n=\mathbf{5{,}000}$ & $n=\mathbf{10{,}000}$& $n=\mathbf{20{,}000}$& $n=\mathbf{30{,}000}$& $n=\mathbf{40{,}000}$ \\
        \midrule 
        \multicolumn{4}{l}{$\pi_{\textsc{theory}}$~\cite{doerr2015black,doerr2018optimal}}&6.723\scriptsize{(0.32)}&6.763\scriptsize{(0.28)}&6.768\scriptsize{(0.26)}&6.834\scriptsize{(0.20)}&6.855\scriptsize{(0.14)}&6.877\scriptsize{(0.13)}&6.882\scriptsize{(0.11)}\\
        \midrule
        \rowcolor{gray!20} $\lambda_m$ & $\alpha$ &$\lambda_c$ & $\beta$ & & & & & & & \\
        \midrule
        Eq.(\ref{eq:handcrafted_lbdm})& 1 & $\lambda_m$ & 1 & 6.353\scriptsize{(0.35)}&6.384\scriptsize{(0.30)}&6.404\scriptsize{(0.28)}&6.475\scriptsize{(0.21)}&6.545\scriptsize{(0.16)}&6.581\scriptsize{(0.15)}&6.602\scriptsize{(0.14)}\\
        $\pi_{\textsc{theory}}$& Eq.(\ref{eq:handcrafted_alpha}) & $\lambda_m$ & 1 &5.905\scriptsize{(0.32)}&5.957\scriptsize{(0.30)}&5.980\scriptsize{(0.26)}&6.044\scriptsize{(0.20)}&6.096\scriptsize{(0.16)}&6.142\scriptsize{(0.15)}&6.157\scriptsize{(0.14)}\\
        $\pi_{\textsc{theory}}$& 1 & Eq.(\ref{eq:handcrafted_lbdc}) & 1 &5.891\scriptsize{(0.24)}&5.923\scriptsize{(0.22)}&5.930\scriptsize{(0.20)}&5.978\scriptsize{(0.15)}&6.029\scriptsize{(0.12)}&6.051\scriptsize{(0.11)}&6.076\scriptsize{(0.10)}\\

        \arrayrulecolor{lightgray}\midrule
        $\pi_{\textsc{theory}}$& Eq.(\ref{eq:handcrafted_alpha}) & Eq.(\ref{eq:handcrafted_lbdc}) & 1 &5.167\scriptsize{(0.24)}&5.199\scriptsize{(0.21)}&5.216\scriptsize{(0.19)}&5.269\scriptsize{(0.14)}&5.309\scriptsize{(0.12)}&5.336\scriptsize{(0.11)}&5.362\scriptsize{(0.10)}\\
        Eq.(\ref{eq:handcrafted_lbdm})& 1 & Eq.(\ref{eq:handcrafted_lbdc}) & 1 &6.006\scriptsize{(0.26)}&6.022\scriptsize{(0.23)}&6.045\scriptsize{(0.20)}&6.090\scriptsize{(0.16)}&6.132\scriptsize{(0.13)}&6.156\scriptsize{(0.11)}&6.172\scriptsize{(0.10)}\\
        \arrayrulecolor{lightgray}\midrule
        Eq.(\ref{eq:handcrafted_lbdm})& Eq.(\ref{eq:handcrafted_alpha}) & Eq.(\ref{eq:handcrafted_lbdc}) & 1 &\textbf{4.827}\scriptsize{(0.23)}&\textbf{4.859}\scriptsize{(0.21)}&\textbf{4.870}\scriptsize{(0.19)}&\textbf{4.930}\scriptsize{(0.15)}&\textbf{4.964}\scriptsize{(0.12)}&\textbf{4.998}\scriptsize{(0.12)}&\textbf{5.006}\scriptsize{(0.10)}\\
        \arrayrulecolor{black}\bottomrule
    \end{tabular}%
    \end{adjustbox}%
\end{table*}%

\textbf{Coefficient $\alpha$ for the mutation rate}. A common trend for this parameter is that it starts jumping to a high value when $f(x) \geq 0.95n$. Therefore, we design a policy where $\alpha$ is set as a small value (e.g., $0.001$) during the first part of the search\footnote{Note that in~\Cref{fig:merged_policies} for a few problem sizes, DDQN appears to suggest a ``U-shape'' policy where $\alpha$ receives a large value before dropping and then increasing. We also experimented with this U-shape policy where we set $\alpha=0.5$ if $f(x) < 0.85n$, $\alpha=0.001$ if $0.85n \leq f(x) \leq 0.95n$, and $\alpha=1$ otherwise. However, this policy performs worse than the one where we simply set $\alpha=0.001$ for all $f(x) \leq 0.95n$. The differences are small, though, this alternative U-shaped policy achieves a normalized ERT of around $6.212$ averaged across seven tested problem sizes.}, and is then increased to $1$ when the fitness reaches a certain distance to the optimum, such as $0.95n$. As a result, the parameter is formulated as follows:

\begin{equation}
\label{eq:handcrafted_alpha}
\alpha =
\begin{cases}
0.001 & \text{if } f(x)/n \le 0.95 \\
1   & \text{otherwise}
\end{cases}
\end{equation}

\textbf{Population size $\lambda_c$ for the crossover phase}. Although the theory-derived policy suggests a dependency of $\lambda_m=\lambda_c$, our RL-based DAC policy imposes a much larger value for $\lambda_c$. The range of this parameter might be approximately twice that suggested by theory. This observation somewhat aligns with the findings in~\cite{dang2019hyper}, where \textsc{irace} also suggested that $\lambda_c = 2\lambda_m$. Notably, the learned DDQN policies are able to recognize this correlation without having to explicitly formulate the relationship between $\lambda_c$ and $\lambda_m$. This allows for more flexibility compared to the \textsc{irace}-based approach: since $\lambda_m$ and $\lambda_c$ are completely decoupled from each other, DDQN can choose to ``delay'' increasing $\lambda_m$ during the early part of the search. The formula for this parameter is as follows: 
\begin{equation}
\label{eq:handcrafted_lbdc}
\lambda_c = 2 \times \sqrt{n/(n-f(x))}
\end{equation}


\textbf{Coefficient $\beta$ for the crossover bias}. The values explored by DDQN for this parameter show a less clear trend, and since this parameter was indicated in~\Cref{sec:importance_ablation_study} as the least important one among the four parameters, we decide to keep this parameter as a constant of $\beta=1$ (i.e., similar to the theory-derived policy).

\textbf{Ablation analysis.} We assess the newly hand-crafted policy across a range of problem sizes from $n=500$ to $n=40{,}000$ and compare it with the theory-derived policy. As shown in the second row ($\pi_{\textsc{theory}}$) and the last row of~\Cref{tab:mpdac_derived_policies}, the derived policy consistently outperforms the theory-derived one across all problem sizes tested, including the very large problem size of $n=40,000$. 

To gain insights into the importance of each component of the derived policy, we conduct an ablation analysis where we replace each of the three parameters, $\lambda_m$, $\alpha$, and $\lambda_c$, with the ones in the theory-derived policy during the ablation path. The results are shown in~\Cref{tab:mpdac_derived_policies} (row 3 onward), where the expected runtime values are again normalized by the problem size. We observe that $\lambda_c$ plays a crucial role in improving the ERT, resulting in an average reduction of over $12\%$, followed by the  mutation rate coefficient $\alpha$ which improves by approximately $11.5\%$. We next examine the combination of two parameters, where the set of $\{\alpha,\lambda_c\}$ exhibits an average improvement of over $22.5\%$, significantly outperforming the set of $\{\lambda_m,\lambda_c\}$ (which only yields a $10\%$ improvement). This observation highlights the negative impact of merging $\lambda_m$ into $\lambda_c$. 
The last row in~\Cref{tab:mpdac_derived_policies} indicates a consistent performance improvement of $27.5\%$ over $\pi_{\textsc{theory}}$, confirming the positive contribution of our hand-crafted policies $\{\lambda_m,\alpha,\lambda_c\}$ to the performance. Consequently, this motivates us to continue exploring fine-grained derived policies for further improvement. 

\subsection{Parameter Tuning for the Newly Derived Policy}

\begin{figure*}[t]
    \centering
    \includegraphics[width=\linewidth, clip]{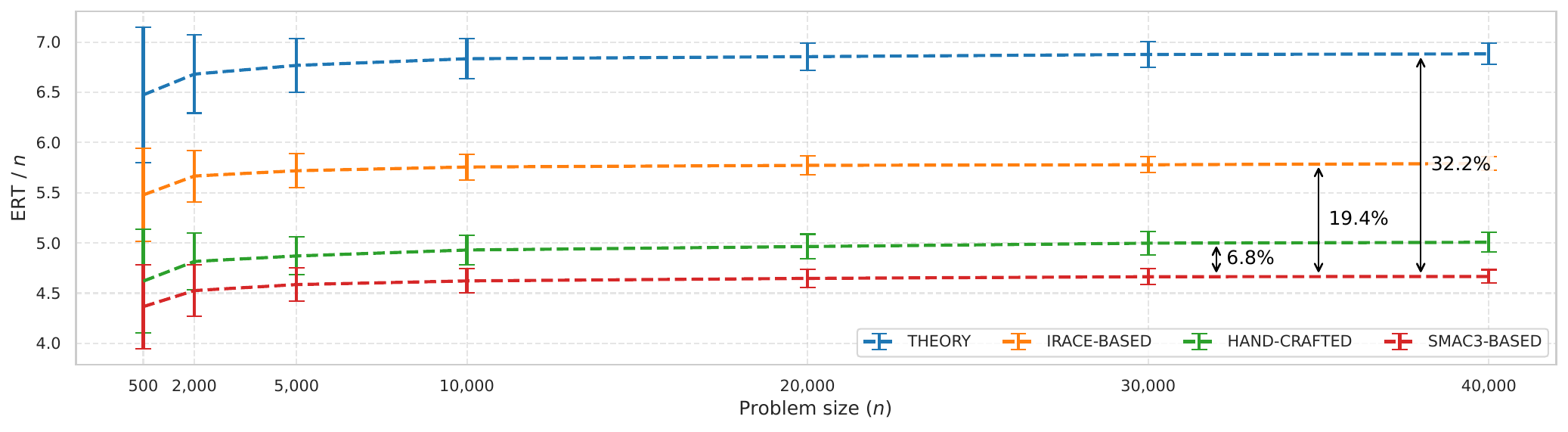}
    \caption{Normalized ERT (and its standard deviation) of our two newly derived policies (\textcolor{ddqn_mp}{\textsc{hand-crafted}} and \textcolor{theory}{\textsc{smac3-based}}), compared to the theory-derived policy~\cite{doerr2015black,doerr2018optimal} (\textcolor{ppo_sp}{\textsc{theory}}) and the \irace-based approach~\cite{dang2019hyper} (\textcolor{ppo_mp}{\textsc{irace-based}}).}
    \vspace{-6pt}
    \label{fig:compare_derived_policy_seen_problem_sizes}
\end{figure*}

\begin{table}[t]
    \centering
    \caption{Comparison of parameters in the newly derived policy with hand-crafted insights and incumbents identified by SMAC3.}
    \label{tab:params_search_space}
    \begin{adjustbox}{max width=\textwidth} 

    \begin{tabular}{clcc}
        \hline
        \textbf{Parameter} & \textbf{Search Space} & \textbf{Hand-crafted} & \textbf{SMAC3} \\
        \hline
        $\kappa_m$ & $\text{interval}(0.5, 1.0)$ & $0.95$ & $0.9596$\\
        $\omega$ & $\text{interval}(1.0, 5.0)$ & $2.0$ & $4.8923$ \\
        $\kappa_\alpha$ & $\text{interval}(0.5, 1.0)$ & $0.95$ & $0.9590$ \\
        $\upsilon$ & $\text{interval}(0.001, 0.1)$ & $0.001$ & $0.00104$ \\
        \hline
    \end{tabular}
    \end{adjustbox}
\end{table}

As demonstrated in the previous sub-section, our insight into the hand-crafted policies is fairly accurate. This encourages us to use the algorithm configuration tool to adjust those policies more finely. Concretely, we re-parameterize the multi-parameter control policy in~\Cref{alg:onell} by introducing four new parameters: (i) the switching point ($\kappa_m$) for $\lambda_m$; (ii) the amplification factor ($\omega$) for $\lambda_c$; (iii) the switching point ($\kappa_\alpha$) and (iv) the lower bound ($\upsilon$) for $\alpha$:
\begin{itemize}
    \item Population size for the mutation phase with $\kappa_m$: 
        \begin{equation}
            \label{eq:new_lbdm}
            \lambda_m =
            \begin{cases}
            1 & \text{if } f(x)/n \le \kappa_m \\
            \sqrt{n/(n-f(x))}   & \text{otherwise}
            \end{cases}
        \end{equation}
    \item Population size for the crossover phase with $\omega$:
    \begin{equation}
    \label{eq:new_lbdc}
    \lambda_c = \omega \times \sqrt{n/(n-f(x))}
    \end{equation}
    
    \item Mutation rate coefficient with $\kappa_\alpha$ and $\upsilon$:
        \begin{equation}
        \label{eq:new_alpha}
        \alpha =
        \begin{cases}
        \upsilon & \text{if } f(x)/n \le \kappa_\alpha \\
        1   & \text{otherwise}
        \end{cases}
        \end{equation}
    \item Crossover bias coefficient: $\beta=1$
\end{itemize}

\textbf{Search spaces.} We define the search spaces for all four parameters: $\kappa_m$, $\omega$, $\kappa_\alpha$, and $\upsilon$ as shown in~\Cref{tab:params_search_space}. While these value ranges are informed by our insights, we adhere to the philosophy that a well-defined search space should balance exploration with computational feasibility.

\textbf{Algorithm configuration setup.} We use the SMAC3 framework~\cite{lindauer2022smac3} as the optimizer. SMAC3 is an upgraded version of SMAC~\cite{hutter2011sequential}, which is a highly established automated algorithm configuration tool.
The optimization employs a multi-fidelity strategy, using the Hyperband~\cite{li2018hyperband} intensifier. The fidelity parameter determines the number of repetitions (budget) for evaluating a single configuration. We allocate 100 repetitions for the initial screening and $1{,}000$ repetitions for high-confidence evaluation. The optimization is limited to a total budget of one million units, which is consumed by the number of repetitions; for instance, a single full-fidelity evaluation uses $1{,}000$ units. This approach ensures that poor configurations are eliminated early (after 100 runs), while promising configurations are accelerated to the maximum fidelity ($1{,}000$ runs) to guarantee statistical significance. In order to speed up the tuning process, we distribute the tuning  across 20 CPUs, allowing for parallel evaluation of candidate configurations. We use the normalized ERT averaged across four problem sizes ($n\in\{200; 500; 1{,}000; 2{,}000\}$) as the objective value for the SMAC3 optimizer.

\begin{figure}[t]
    \centering
    \includegraphics[width=0.8\linewidth, trim={0 5.5cm 0 0}, clip]{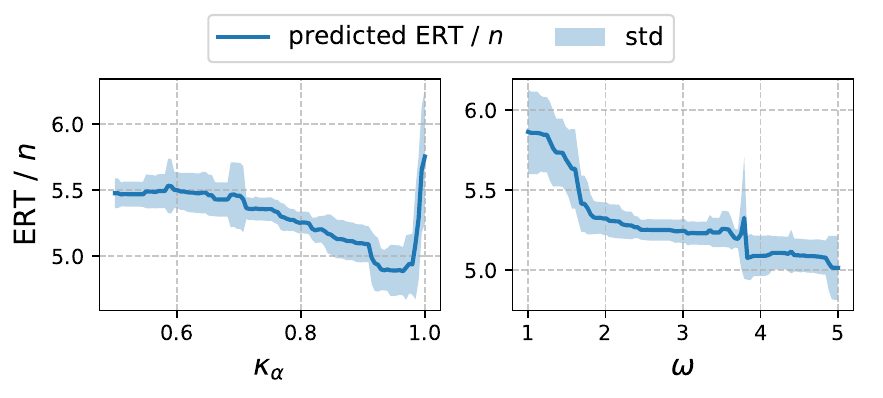}
    \vspace{-6pt}
    \caption{Marginal performance plots of $\kappa_\alpha$ and $\omega$, the two most important parameters during SMAC3's tuning process, identified by fANOVA. The y-axis shows the marginal performance, i.e., the average algorithm performance predicted by fANOVA for a given parameter value (x-axis).}
    \label{fig:f_anova}
    \vspace{-6pt}
\end{figure}

The incumbents identified by SMAC3, as shown in~\Cref{tab:params_search_space}, are notably very close to our hand-crafted values. We provide a comprehensive performance evaluation across four methods: the theory-derived policy ($\pi_{\textsc{theory}}$~\cite{doerr2015black,doerr2018optimal}), the strongest baseline achieved by \textsc{irace} in previous study, and our two newly derived policies described above. As illustrated in~\Cref{fig:compare_derived_policy_seen_problem_sizes}, our policy supported by SMAC3 significantly surpasses the others for all tested problem sizes. Notably, the normalized ERT seems to remain constant across all problem sizes, approximately $6.9$, $5.8$, $5.0$, and $4.7$ for the theory-derived policy, the \textsc{irace}-based one, and our two newly developed policies, respectively, suggesting a generalizable constant speedup that could be studied in future theoretical analysis on this benchmark.

As shown in~\Cref{tab:params_search_space}, after tuning with SMAC3, parameter $\upsilon$ remains unchanged, while $\kappa_m$ and $\kappa_\alpha$ stay close to their default values. Only $\omega$ undergoes a substantial change. To better understand the significance of the decisions made by SMAC3, we employ the parameter importance analysis tool fANOVA~\cite{hutter2014efficient}. Developed by the same authors as SMAC, fANOVA reuses the performance data collected during the tuning process to provide insights into the configurator’s decisions. Specifically, fANOVA constructs a random forest model that takes an algorithm configuration as input and predicts the corresponding performance. This model is then used to quantify the contribution of each parameter, as well as their low-order interactions, to the variance in performance.

Interestingly, fANOVA identifies the switching point $\kappa_\alpha$ and the amplification factor $\omega$ as the two most important parameters, accounting for $30.84\%$ and $30.29\%$ of the total performance variance, respectively. These findings suggest that (i) our hand-crafted rule for $\lambda_m$ is already close to optimal; and (ii) it is crucial to decouple the two population sizes, $\lambda_m$ and $\lambda_c$, with a substantially larger $\lambda_c$ being highly beneficial. 

The remaining parameters, namely the switching point $\kappa_m$ and the lower bound $\upsilon$, have considerably smaller effects, contributing $7.25\%$ and $2.52\%$ to the total variance, respectively. There is also a $7.79\%$ pairwise interaction effect between the two most important parameters, ${\kappa_\alpha, \omega}$, followed by a $6.91\%$ interaction between ${\kappa_m, \omega}$. These interaction effects indicate that the corresponding parameter pairs cannot be tuned independently.

\section{Conclusion}

In this work, we have shown that (deep) RL can be a powerful tool to discover high-performing,  theoretically tractable multi-parameter control policies for evolutionary computation. By leveraging results suggested by deep-RL for controlling four parameters of the the \onell optimizing \onemax, and then finetuning these policies with an algorithm configuration tool, we have derived a symbolic policy that outperforms all existing policies on this task. 

We believe our work offers new insights for researchers interested in reinforcement learning, algorithm configuration, or in theoretical aspects of evolutionary computation (EC). For the EC community, it is worth noting that our derived policies are fairly simple and may be tractable for a fine-grained running time analysis. However, we note that, to date, no reasonably good bounds for the constant factors of the linear expected running time of the \onell with suitable dynamic parameter choices are known, even though experimental results for up to $n=2^{22}$ in~\cite{AntipovBD22} suggest that the constant factor of the default 1/5-th success rule proposed in~\cite{doerr2015black,doerr2018optimal} is somewhere around $6.9$, perfectly in line with our empirical results. The interested reader may have noticed that the constant factors for all algorithms and parameter control policies increase with increasing problem dimension $n$, and this includes the settings for which we have proven guarantees of linear running times. However, as~\cite[Section~4.3]{AntipovBD22} acknowledges, the few formally proven upper bounds for the constant factors (for their algorithm choosing $\lambda$ from a heavy-tailed power-law distribution) seem to be far too pessimistic. Since the precision of running time results has drastically increased over the last years, fine-grained performance analyses of the original self-adjusting \onell and the \onell equipped with our discovered policies may already be feasible or become feasible in the near future. 

From a more general perspective, we hope that this work, in line with others e.g., ~\cite{VermettenLRBD24} (expanding upon~\cite{LenglerR22}) and~\cite{DoerrDL21} (deriving running time guarantees for observations made in~\cite{DoerrW18}) underlines the benefits of tying sound empirical investigations with rigorous theoretical analyses, and vice-versa.  


On the empirical side, we conjecture that the more robust deep-RL policies provide a richer foundation for symbolic policy discovery. Therefore, further improving the performance of deep-RL algorithms is one of the most feasible next steps. Multi-agent reinforcement learning (MARL)~\cite{sunehag2017value,rashid2020monotonic,yu2022surprising} can serve as a guiding principle, especially given its recent examination in the DAC context~\cite{xue2022multi,lu2025sequential}. Concerning the failures of policy-gradient-based RL methods, we plan to explore the long-standing concept of offline RL~\cite{levine2020offline} and Imitation Learning (IL)~\cite{torabi2018behavioral}, where we anticipate that offline transitions will assist RL algorithms in exploring effective policies. For policy discovery, several straightforward approaches, such as the tree search~\cite{de2018greedy} 
can be used to approximate the RL agent's policy. Another advanced technique is to simultaneously conduct the RL training and symbolic policy regression~\cite{delfosse2023interpretable,gu2024pi}. 

\section*{Acknowledgments}
The project is financially supported by the European Union (ERC, ``dynaBBO'', grant no.~101125586), by ANR project ANR-23-CE23-0035 Opt4DAC, and by an International Emerging Action funded by CNRS Sciences informatiques. This work used the supercomputer at MeSU Platform (\href{https://sacado.sorbonne-universite.fr/plateforme-mesu}{https://sacado.sorbonne-universite.fr/plateforme-mesu}). Tai Nguyen acknowledges funding from the St Andrews Global Doctoral Scholarship programme. This publication is based upon work from COST Action CA22137 ``Randomized Optimization Algorithms Research Network'' (ROAR-NET), supported by COST (European Cooperation in Science and Technology).

\bibliographystyle{IEEEtran_minimal}
\bibliography{refs}


\end{document}